%% file: main.tex
\pdfoutput=1

\documentclass[11pt]{article}

\usepackage{setspace}
\PassOptionsToPackage{hyphens}{url}

\usepackage{assets/acl}
\input{preamble}

\def\papertitle{\textsf{AnchorAL}: Computationally Efficient Active Learning for Large and Imbalanced Datasets}

\title{\papertitle}

\author{%
  Pietro Lesci \& Andreas Vlachos\\
  Department of Computer Science and Technology \\
  University of Cambridge \\ 
  \myemph{\{}\href{mailto:pl487@cam.ac.uk}{\myemph{pl487}}, \href{mailto:av308@cam.ac.uk}{\myemph{av308}}\myemph{\}@cam.ac.uk}
}

\begin{document}

% ==== Title and authors
\maketitle

% ==== Main body
\input{body}

% ==== Bibliography
\bibliography{biblio.bib}

% ==== Appendix
\clearpage
\appendix
\input{appendix}

\end{document}

%% file: preamble.tex
\usepackage{adjustbox}
\usepackage{multirow}
\usepackage{wrapfig}
\usepackage{scalerel}
\usepackage{float}
\usepackage[caption = false]{subfig}

\usepackage[T1]{fontenc}    % use 8-bit T1 fonts

\usepackage[utf8]{inputenc}

\usepackage{microtype}

\usepackage{inconsolata}

\usepackage{xltabular}
\usepackage{lscape} % for landscape orientation
\usepackage{times}
\usepackage{latexsym}
\usepackage{datetime}       % dates
\usepackage{booktabs}       % professional-quality tables
\usepackage{nicefrac}       % compact symbols for 1/2, etc.
\usepackage{xcolor}         % colors
\usepackage{colortbl}
\usepackage{url}            % simple URL typesetting
\usepackage{fnpct}           % nicer footnotes
\usepackage{enumitem}        % nicer itemize
\usepackage{tcolorbox}       % colored background
\usepackage{longtable}
\usepackage{csquotes}        % quotation marks
\usepackage{xspace}          % adds whitespace after macros

\makeatletter
\xspaceaddexceptions{\csq@qclose@i}
\makeatletter

\usepackage[scaled=0.85]{helvet}

\usepackage{bbm}
\usepackage{amsfonts}
\usepackage{amssymb}
\usepackage{mathtools}
\usepackage{amsmath}
\usepackage{bm}
\usepackage{amsthm}

\DeclareMathOperator*{\argmax}{argmax}

\usepackage{algorithm}
\usepackage{algpseudocode}  % requires algorithmicx
\makeatletter

\algnewcommand{\LineComment}[1]{\Statex \hskip\ALG@thistlm \textcolor{blue}{\(\triangleright\) #1}}

\algnewcommand{\FirstLineComment}[1]{\Statex \hskip\ALG@tlm \textcolor{blue}{\(\triangleright\) #1}}

\algnewcommand{\InlineComment}[1]{\hfill\textcolor{blue}{\(\triangleright\) #1}}
\makeatother

\usepackage{cleveref}
\crefname{section}{\S}{\S\S}
\Crefname{section}{\S}{\S\S}
\crefformat{section}{\S#2#1#3}
\crefname{figure}{Fig.}{Fig.}
\crefname{algorithm}{Alg.}{Alg.}
\crefname{line}{line}{lines}
\crefname{appendix}{App.}{}
\crefname{equation}{eq.}{eqs.}
\crefname{table}{Table}{Tables}
\crefname{proposition}{Proposition}{Propositions}
\crefname{assumption}{Assump.}{Assumps.}
\crefname{definition}{Definition}{Definitions}

\usepackage{tabularray}
\usepackage{cellspace}
\newcommand\cincludegraphics[2][]{\raisebox{-0.3\height}{\includegraphics[#1]{#2}}}
\usepackage{setspace}

\usepackage{siunitx}
\newcommand{\q}[2]{\qty[mode=math]{#1}{#2}\xspace}

\DeclareSIUnit[quantity-product = {}, reset-math-version = false]\thousand{k}
\DeclareSIUnit[quantity-product = {}, reset-math-version = false]\million{M}
\DeclareSIUnit[quantity-product = {}, reset-math-version = false]\billion{B}
\DeclareSIUnit[quantity-product = {}, reset-math-version = false]\trillion{T}

\DeclareSIUnit[quantity-product = {}, reset-math-version = false]\x{x}
\DeclareSIUnit[quantity-product = {}, reset-math-version = false]\percent{\%}

\DeclareSIUnit[quantity-product = {}, reset-math-version = false]\hour{h}
\DeclareSIUnit[quantity-product = {}, reset-math-version = false]\min{m}
\DeclareSIUnit[quantity-product = {}, reset-math-version = false]\sec{s}

\newcommand{\integer}[1]{%
  \num[
    mode = math,
    round-mode=places,
    round-precision=0,
    group-separator={,},
    group-minimum-digits=4,
  ]{#1}%
  \xspace
}

\newcommand{\float}[2][1]{%
  \num[
    group-digits=false,
    round-precision=#1,
    round-mode=places,
  ]{#2}%
  \xspace}

\newcommand{\percentage}[2][1]{%
  \qty[
    round-mode=places,
    round-precision=#1,
    output-decimal-marker={.},
    scientific-notation = fixed,
  ]{#2e2}{\percent}%
  \xspace
}

\newcommand{\snum}[1]{\num{#1}\xspace}

\xspaceaddexceptions{\textsubscript}

\makeatletter
\DeclareRobustCommand*{\escapeus}[1]{%
  \begingroup\@activeus\scantokens{#1 }\endgroup}
\begingroup\lccode`\~=`\_\relax
\lowercase{\endgroup\def\@activeus{\catcode`\_=\active \let~\_}}
\makeatother

\newcommand{\prettyurl}[1]{\fontsize{10pt}{10pt}\url{#1}}

\newcommand{\mydef}[2]{\def#1{#2\xspace}}

\newcommand{\mathdef}[2]{\expandafter\newcommand\csname #1\endcsname{\ensuremath{#2}}}

\newcommand{\format}[1]{{\sffamily#1}}
\newcommand{\myemph}[1]{\textsf{{\escapeus{#1}}}}

\newlength{\maxfontsize}
\newcommand{\code}[1]{%
  \ifdim\fontdimen6\font>\maxfontsize
    {\fontsize{8.5pt}{10pt}\format{\textbf{\escapeus{#1}}}}%
  \else
    \format{\textbf{\escapeus{#1}}}%
  \fi
}

\newcommand{\size}[1]{{\vert #1 \vert}}
\newcommand{\makebig}[3][1]{%
  \raisebox{#1}{\scalebox{#2}{$\mathbf{#3}$}}%
}

\mydef{\AL}{AL}

\mathdef{inspace}{\mathcal{X}}
\mathdef{outspace}{\mathcal{Y}}
\mathdef{embspace}{\mathcal{E}}
\mathdef{naturals}{\mathbb{N}}
\mathdef{reals}{\mathbb{R}}
\mathdef{pool}{\mathcal{U}}
\mathdef{subpool}{\widetilde{\pool}}
\mathdef{dataset}{\mathcal{D}}
\mathdef{queryset}{\mathcal{B}}
\mathdef{anchorset}{\mathcal{A}}
\mathdef{modelset}{\mathcal{F}}
\mathdef{minorityset}{\mathcal{C}}
\mathdef{nnset}{\mathcal{N}}
\mathdef{avgnnset}{\widetilde{\nnset}}
\mathdef{simscores}{\mathbf{s}}
\mathdef{simscore}{s}
\mathdef{myindex}{\mathcal{I}}

\mathdef{strategy}{\Phi}
\mathdef{model}{f}
\mathdef{knn}{\mathtt{kNN}}
\mathdef{topk}{\mathtt{argmax}}
\mathdef{anchorstrategy}{\Gamma}
\mathdef{aggfn}{\mathtt{avg}}
\mathdef{trainfn}{\mathtt{train}}
\mathdef{encoder}{g}
\mathdef{simfn}{\Delta}

\def\cerchio{\makebig[-1.5pt]{1.7}{\circ}}
\def\cross{\makebig[-.5pt]{1.2}{\times}}
\def\point{\makebig[-.5pt]{1.2}{\bullet}}
\mathdef{subpoolsize}{\bar{M}}
\mathdef{maxsubpoolsize}{M}
\mathdef{querysize}{b}
\mathdef{numneighbours}{K}
\mathdef{numanchors}{A}
\mathdef{oracle}{f^\star}
\mathdef{numclasses}{C}
\mathdef{class}{c}
\mathdef{numrounds}{T}
\mathdef{poolx}{\vx_u}
\mathdef{budget}{B}
\mathdef{prob}{\mathbb{P}}
\mathdef{vx}{\boldsymbol{x}}

\mydef{\agnewsbus}{\myemph{Agnews-Bus}}
\mydef{\agnews}{\myemph{Agnews}}
\mydef{\amazonagri}{\myemph{Amazon-Agri}}
\mydef{\amazonmulti}{\myemph{Amazon-Multi}}
\mydef{\wikitoxic}{\myemph{WikiToxic}}
\mydef{\Agnews}{Agnews}
\mydef{\Amazoncat}{AmazonCat-13k}
\mydef{\Wikitoxic}{WikiToxic}

\mydef{\kmeanspp}{\textit{k}-\textsc{means++}}
\mydef{\kmeans}{\textit{k}-\textsc{means}}
\mydef{\name}{\myemph{AnchorAL}}
\mydef{\random}{\myemph{Random}}
\mydef{\randomsubset}{\myemph{RandSub}}
\mydef{\seals}{\myemph{SEALS}}
\mydef{\noop}{\myemph{No-Op}}
\mydef{\entropy}{\myemph{Entropy}}
\mydef{\ftbertkm}{\myemph{FT-BERTKM}}
\mydef{\badge}{\myemph{BADGE}}
\mydef{\smallbudget}{\myemph{small}}
\mydef{\mediumbudget}{\myemph{medium}}
\mydef{\largebudget}{\myemph{large}}
\mydef{\tinymodel}{\myemph{tiny}}
\mydef{\basemodel}{\myemph{base}}

\mydef{\bertbase}{\myemph{BERT-base}}
\mydef{\albertbase}{\myemph{ALBERT-base}}
\mydef{\berttiny}{\myemph{BERT-tiny}}
\mydef{\debertabase}{\myemph{DeBERTa-base}}
\mydef{\gpt}{\myemph{GPT-2}}
\mydef{\tf}{\myemph{T5-base}}

%% file: body.tex
% ========
% ABSTRACT
% ========
\begin{abstract}
    Active learning for imbalanced classification tasks is challenging as the minority classes naturally occur rarely. Gathering a large pool of unlabelled data is thus essential to capture minority instances. Standard pool-based active learning is computationally expensive on large pools and often reaches low accuracy by overfitting the initial decision boundary, thus failing to explore the input space and find minority instances.
    To address these issues we propose \name. At each iteration, \name chooses class-specific instances from the labelled set, or \emph{anchors}, and retrieves the most similar unlabelled instances from the pool. This resulting \emph{subpool} is then used for active learning.
    Using a small, fixed-sized subpool \name allows scaling any active learning strategy to large pools. By dynamically selecting different anchors at each iteration it promotes class balance and prevents overfitting the initial decision boundary, thus promoting the discovery of new clusters of minority instances.
    In experiments across different classification tasks, active learning strategies, and model architectures \name is \textit{(i)} faster, often reducing runtime from hours to minutes, \textit{(ii)} trains more performant models, \textit{(iii)} and returns more balanced datasets than competing methods.
\end{abstract}

% ===========
% GITHUB LINK
% ===========
%\noindent
\begin{tblr}{colspec = {Q[c,m]|X[l,m]}, stretch = 0}
    \cincludegraphics[width=1.2em, keepaspectratio]{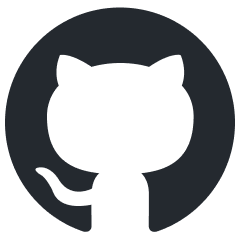} & \setstretch{.5}\href{https://github.com/pietrolesci/anchoral}{\myemph{pietrolesci/anchoral}} \\
\end{tblr}

% ============
% INTRODUCTION
% ============
\section{Introduction}\label{sec:introduction}

The abundance of web-scale textual data \footnote{E.g.\ \href{https://commoncrawl.org}{Common Crawl corpus} is in the order of petabytes.} has contributed to the success of generalist language models pretrained as multi-purpose foundation models and fine-tuned to solve downstream natural language processing tasks \citep{bommasani-etal-2022-opportunities}. The data used during fine-tuning critically affects their downstream abilities \citep{lee_deduplicating_2022, gururangan_whose_2022}, especially in real-world applications where performance on rare \emph{concepts}, or minority classes, is critical \citep{he_learning_2009}.

\input{assets/images/method}

\input{assets/images/initial_plot}

Data collection and annotation for imbalanced classification tasks is challenging as the minority class(es) naturally occur(s) rarely. Gathering a large pool of unlabelled data is often essential to capture minority instances making manual annotation prohibitively expensive. In principle, considering only a random subset of the pool is potentially amenable to manual curation but can be suboptimal when the imbalance is large as instances of the rare class might never be found. Therefore, the learning challenges caused by the imbalance and the computational challenges stemming from considering a large pool are often intrinsically intertwined and need to be jointly addressed.
Active Learning (\AL) provides an automatic mechanism to prioritise instances by allowing a model to select those to label, resulting in a higher label efficiency and lower annotation costs \citep{hanneke_bound_2007, hanneke_rates_2011, balcan_agnostic_2009, balcan_true_2010}. Also, it can partially mitigate mild imbalance \citep{ertekin_learning_2007}.

Standard pool-based active learning (\AL), however, struggles with large and imbalanced pools.
First, \AL can be too computationally expensive due to its iterative nature. Specifically, the inference costs of repeatedly evaluating a model on every unlabelled instance in each iteration can be prohibitive, especially given the size of modern language models \citep{tsvigun_towards_2022}; furthermore, the annotation costs (i.e., the man-hours per annotation) can drastically increase due to the annotators' waiting time between iterations.
Second, \AL can be as good as random selection due to the imbalance: diversity-based strategies can be ineffective when minority and majority instances are not easily separable in high-dimensional spaces \citep{thudumu_comprehensive_2020}, while uncertainty-based strategies can fail to explore the input space and discover minority clusters because they tend to keep refining the initial decision boundary \citep{tomanek_proper_2009}; hybrid strategies suffer from similar limitations.
One line of prior work has proposed to subset the pool in each iteration before running an \AL strategy to speed up instance selection \citep{ertekin_learning_2007, coleman_similarity_2022}. However, existing pool filtering methods either do not address class imbalance or are computationally inefficient. Therefore, it is still an open question how to scale \AL on large datasets while addressing the imbalance.

In this paper, we propose a new method, \name, designed to jointly address the learning and computational challenges of applying \AL to large and imbalanced datasets. \name works by filtering the pool before running any \AL strategy, thus reducing the inference costs and the annotators' waiting time in each iteration. Crucially, the filtering process is designed to promote the exploration of the input space and the discovery of minority instances while keeping instance selection time constant, regardless of the original pool size.
Specifically, at each iteration, \name chooses class-specific instances from the labelled set, which we term \emph{anchors}. Then, each unlabelled instance is scored based on its average distance from the anchors and the most similar are used to form the subset of the pool, which we term \emph{subpool}.
While any similarity measure works, we use the semantic representation capabilities of language models (e.g., \citealp{devlin_bert_2019}) and measure similarity based on cosine distance between instance representations.
Using a small, fixed-sized subpool \name allows scaling any active learning strategy to large pools while keeping the annotators' waiting time constant across iterations independently of the size of the original pool. Optionally, the maximum size of the subpool can be set by the user, thus making \name suitable for different computational budgets.
Moreover, by dynamically selecting different anchors at each iteration it prevents overfitting the initial decision boundary, thus promoting the discovery of new clusters of minority instances; the class-specific anchors promote class balance even for \AL strategies that do not explicitly account for imbalance (\Cref{fig:glance}).

We test the effectiveness of \name across \integer{4} text classification tasks, both binary and multiclass, \integer{3} \AL strategies (i.e., uncertainty, diversity, and hybrid), and \integer{6} models with different sizes and architectures (i.e., encoder, decoder, and encoder-decoder).
Experiments show that \name is the fastest method, reducing the total selection time from hours to minutes; (often) the best-performing, reaching higher performance in less time and with fewer annotations; and discovers the most minority instances resulting in more balanced labelled sets.

% ============
% RELATED WORK
% ============
\section{Related Work}\label{sec:related_work}

In this section, we provide the background on imbalanced learning and present the challenges of applying \AL to imbalanced datasets. Then, we discuss approaches to address these challenges and scale \AL to large pools.

\subsection{Learning from Imbalanced Datasets}

The research on imbalanced learning \citep{johnson_survey_2019, henning_survey_2023} can be broadly divided into two approaches. Data-level approaches directly balance the training data distributions by generating synthetic samples \citep{mullick_generative_2019} or resampling the available data \citep{chawla_smote_2002, van_hulse_experimental_2007}. Re-weighting approaches assign a different weight to each instance to adjust its contribution to the training loss and thus its importance in determining the parameter updates based on the stochastic gradients \citep{dong_class_2017, wang_learning_2017, lin_focal_2017, cui_class-balanced_2019, park_influence-balanced_2021}.

Although \name can be associated with data-level approaches for its re-balancing effect, it differs from those in two crucial aspects. First, the objective of \name is to choose \emph{which} data to learn from rather than optimise \emph{how} to learn from data. Second, \name targets a setting in which the labels are not known ex-ante, that is \AL. Therefore, it cannot easily rely on data augmentation, generation, or over/under-sampling. Instead, it uses the semantic representation capabilities of language models and smartly selects anchors to use as queries to retrieve a more balanced subpool. In \Cref{tab:ablations} we show that \textit{which} and \textit{how many} anchors are chosen strongly affects performance.

\subsection{Class Imbalance in Active Learning}

Standard pool-based \AL, by design, tends to explore the regions nearest to the current decision boundary to \emph{refine} the learned decision boundary until it eventually approaches the optimal one \citep{osugi_balancing_2005}. However, the initial approximation can be distorted and converge to a suboptimal decision boundary when the data is imbalanced. Specifically, the class distribution derived from the initial imbalanced labelled set results in a rough initial decision boundary that makes the model overconfident about predicting the majority class everywhere as the low-certainty regions are just around the minority instances initially seen \citep{park_influence-balanced_2021}. As a result, the \AL strategy fails to explore the input space (\Cref{fig:method}) and keeps refining the known decision boundary \citep{baram_online_2004, dligach_good_2011}, a phenomenon that we term \emph{path dependence}.
Path dependence is exacerbated when minority instances sparsely occupy the input space instead of forming tight clusters. If the initial set contains minority instances that are only from a specific cluster the \AL strategy can even exhaust its budget before discovering new clusters \citep{attenberg_why_2010, attenberg_class_2013}. This issue has long been studied in \AL and historically known as \emph{missed cluster effect} \citep{schutze_performance_2006} or \emph{hasty generalisation} \citep{wallace_active_2010}.\looseness=-1

Addressing path dependence and the missed cluster effect is especially relevant in the context of language models fine-tuning because recent large pretrained models have the potential to memorise their entire training data \citep{kim_bag_2023} and tend to overfit to the overlapping regions between classes \citep{arpit_closer_2017, zhang_understanding_2021}.

\subsection{Actively Learning Imbalanced Datasets}

To address class imbalance in \AL, prior work proposes to use new \AL strategies \citep[\emph{inter alia}]{aggarwal_active_2020, kothawade-etal-2021-similar,jin-etal-2022-deep,zhang-etal-2022-galaxy,zhang-etal-2023-algorithm}, change the interaction protocol by allowing annotators to search for instances to label \citep{attenberg_why_2010, balcan_robust_2012}, or optimise for the recall of minority instances rather than model performance \citep{garnett_bayesian_2012, jiang_efficient_2018, jiang_cost_2019}.
These approaches dictate specific \AL strategies or drastically differ from standard \AL and, thus, unlike AnchorAL, cannot be combined with arbitrary AL strategies, and are out of scope. Moreover, although search has been shown to theoretically improve label efficiency \citep{beygelzimer_search_2016}, in practice it is significantly more expensive since the annotation process is more challenging and thus requires more time. Also, search can be suboptimal beyond the initial data collection phase \citep{levonian_trade-offs_2022} and crucially depends on the annotators' ability to find useful keywords for the domain of interest \citep{hartford_exemplar_2020}.%, and might not be feasible in some cases: for example it might not be possible to search for a specific hateful \emph{sense} of a word \citep{hartford_exemplar_2020}. 

Instead, our work finds its roots in \citet{ertekin_learning_2007} that originally proposed to randomly sample a fixed-sized subset of the pool at each iteration to speed up the \AL process, an approach that we refer to as \randomsubset.
While motivated by its purely computational benefits, our intuition is that pool filtering can address the imbalance too: choosing a different subset of the pool in each iteration forces the \AL strategy to explore different parts of the input space instead of focusing on refining the initially learned decision boundary. Intuitively, a more extensive exploration of the input space promotes the discovery of new minority instances.
Recently, \citet{coleman_similarity_2022} showed that random subsampling can be ineffective when the imbalance is high and proposed \seals which restricts the pool to the $k$-nearest neighbours of all labelled instances. However, \seals suffers from computational inefficiencies because the size of the subpool grows throughout the \AL process: when a new instance is labelled its $k$ neighbours are added to the subpool and are not removed unless labelled. Also, when the initial labelled set is small, the resulting initial subpool is small too (when $k$ is small) or redundant (when $k$ is large) which makes subsequent subpools similar to each other thus limiting the exploration of the input space due to the path dependence.

\name shares the same motivation as \randomsubset and \seals but proposes an alternative approach that allows keeping the subpool size fixed across iterations while still effectively promoting the exploration of the input space, thus combining the benefits of both without none of the downsides.
Similarly to \randomsubset, \name chooses a different, fixed-sized subpool at each iteration by selecting different anchors which makes it \emph{dynamically} adapt to the changing labelled set during \AL. Instead, for \seals the subpool size grows by $k$ units per each new labelled instance and after a few iterations the subpool remains almost unchanged because the new instances are only a small fraction of the entire subpool.
Similarly to \seals, \name relies on similarity search to discover minority instances while \randomsubset can fail under extreme class imbalance. However, differently from \seals, \name emphasises query selection (i.e., anchors) as it targets retrieval of useful instances rather than purely maximising recall of minority instances.

\subsection{Active Learning on Large Pools}

The computational overhead required by the iterative nature of pool-based \AL when applied to large pools, especially in combination with large models, can make \AL infeasible. Recent work in computationally efficient \AL has proposed to use smaller models as cheap proxies \citep[\emph{inter alia}]{yoo_learning_2019, coleman_selection_2020}, selecting large batches that are both informative and diverse to reduce the number of labelling iterations necessary to reach a target performance \citep[\emph{inter alia}]{sener_active_2018, kirsch_batchbald_2019,pinsler_bayesian_2019}, or generating examples \citep[\emph{inter alia}]{lin_active_2018,mayer_adversarial_2020}. However, proxies and large-batch approaches still require evaluation over the entire pool, and generative methods struggle to match the label efficiency of traditional \AL \citep{settles_active_2012}.
Instead, \name directly limits the size of the pool. Besides the rebalancing effects discussed in the previous sections, this approach allows scaling up to large pools any \AL strategies and models without modifications.

\begin{tcolorbox}[title={Advantages of \name vs Prior Work}]
    \name promotes the exploration of the input space by dynamically selecting different anchors which then retrieve a different subpool at each iteration, thus preventing the \AL strategy from overfitting the initial labelled set. Moreover, by keeping the size of the subpool fixed and small across iterations, it can scale to large pools independently of their original size.
\end{tcolorbox}

% ===========
% METHODOLOGY
% ===========
\section{Methodology}\label{sec:methodology}

\input{assets/alg_anchoral}

We consider the standard pool-based \AL setting for classification tasks \citep{atlas_training_1989} wherein we are given a (large) pool of unlabelled data $\pool$ and access to an oracle $\oracle$ that given an instance $\vx$ returns its true\footnote{We do not consider noisy oracles \citep{settles_active_2012}.} label $y \in \{1, ..., C\}$. The goal is to induce the best possible classifier $\model$ within a fixed annotation budget $\budget$.
The annotation process happens iteratively over $\numrounds$ iterations. We assume access to a small initial labelled set $\dataset_0$ with at least one instance per class used to bootstrap an initial classifier $\model_0$.
In each iteration, we train a new version of the classifier $\model_t$ on the available labelled instances $\dataset_t$. Then, the \AL strategy $\strategy$ uses the model to inform the selection of the set of instances to annotate $\queryset_t \subset \pool_t$ such that $\size{\queryset_t} \mathop{=} \lfloor\budget / \numrounds\rfloor$. Finally, the selected instances are annotated by the oracle, removed from the pool, and added to the labelled set; and the cycle repeats.

\subsection{\textsf{AnchorAL}: Anchored Pool Filtering}

\name runs the \AL strategy only on a subset of the pool $\subpool \subset \pool$, such that $\size{\subpool}\mathop{\leq}\maxsubpoolsize \mathop{\ll} \size{\pool}$ where $\maxsubpoolsize$ is a fixed, user-defined upper bound on the subpool size.
To construct the subpool $\subpool$ in a way that promotes exploration and favours minority cluster discovery beyond those present in the initial set, \name anchors the filtering process to a set of labelled instances $\anchorset_t\subset\dataset_t$ created selecting $\numanchors$ instances from each class, such that $\size{\anchorset_t} \mathop{=} \numanchors \mathop{\times} \numclasses$. These anchors are dynamically selected and differ at each iteration. Then, the unlabelled instances are scored based on their average distance from the anchors and the $\subpoolsize\leq\maxsubpoolsize$ most similar are used as subpool (\Cref{alg:algorithm}). In the following paragraph we detail these two parts of the algorithm.

\begin{tcolorbox}[title={Intuition}]
    \name \enquote{biases} the subpool towards smaller regions of the input space so that the resulting subpool only represents these regions and forces the \AL strategy to focus on those. Since anchors are chosen according to a diversity criterion, in each iteration the resulting subpool represents a small and different region of the input space, thus promoting exploration and consequently selection of minority instances for labelling.
\end{tcolorbox}

\paragraph{Anchor Selection (\Crefrange{alg:anchorselect_start}{alg:anchorselect_end}).}
The anchor selection strategy $\anchorstrategy$  can be class-specific but using only one strategy for all classes is enough in practice. Similarly, we select the same number of anchors $\numanchors\mathop{=}\integer{10}$ per each class. To elicit the discovery of new minority clusters, we choose an anchor selection strategy that promotes diversity and use the \kmeanspp initialisation scheme \citep{arthur_k-means_2007}\footnote{As implemented in \href{https://scikit-learn.org/stable/modules/generated/sklearn.cluster.kmeans_plusplus.html}{\myemph{sklearn.cluster.kmeans\_plusplus}}.}. It was originally proposed to produce a good initialisation for \kmeans clustering \citep{steinhaus_sur_1956} and works by iteratively sampling points in proportion to their squared distances from the nearest points already chosen. We run it for each class separately and apply it to the sentence representations derived as described in the next paragraph. We discuss other strategies in \Cref{sec:analysis}.

\paragraph{Similarity Scoring (\Crefrange{alg:knn}{alg:avg}).}
To score unlabelled instances we need to define a similarity measure. While any similarity measure works (e.g., BM25 by \citealp{robertson-zaragoza-2009-bm25}) we use the semantic representation capabilities of language models and measure similarity based on cosine distance between instance representations. Thus, we construct a dense index $\myindex$, that is kept fixed for the entire \AL process, using a pre-trained encoder. Specifically, we use MPNet \citep{song_mpnet_2020}, which is trained to use cosine distance as the similarity function, to encode all available instances (i.e., $\pool_0\cup\dataset_0$).
Generating the embeddings can be computationally expensive but it is performed once and its cost is amortised over the entire \AL process. Also, often the encoder has a similar or smaller size than the model we train, thus creating the embeddings has approximately the same cost as running one iteration of standard \AL.\footnote{Encoding can be sped up using efficient procedures (\href{https://github.com/huggingface/optimum-benchmark/tree/57d69c61be892640d1eb665cc1fe6ab0d8f39c4e/examples/fast-mteb}{\myemph{optimum/fast-mteb}}) and new encoders (\href{https://huggingface.co/spaces/mteb/leaderboard}{\myemph{mteb/leaderboard}}).}
Given the embeddings, we create a searchable index $\myindex$ using the Hierarchical Navigable Small World algorithm \citep{malkov_efficient_2018}\footnote{As implemented in the \href{https://github.com/nmslib/hnswlib}{\myemph{hnswlib}} library (\Cref{app:index_construction}).}, an approximate nearest-neighbour search method that can easily scale to extremely large (\q{>1}{\billion}) datasets with retrieval times in the milliseconds \citep{johnson_billion-scale_2017,babenko_efficient_2016}.

To make the scoring mechanism efficient, instead of scoring each unlabelled instance against each anchor, we retrieve the $\numneighbours = \integer{50}$ nearest neighbours of each anchor from the pool $\nnset_t \subset \pool_t$ (\Cref{alg:knn}) and the relative similarity scores such that $\size{\nnset_t}\leq\numanchors\mathop{\times}\numclasses\mathop{\times}\numneighbours$, where the equality is not strict because in practice different anchors can retrieve the same unlabelled instance. When duplicates are retrieved, we average their similarity scores and get a de-duplicated set of neighbours, $\avgnnset$ (\Cref{alg:avg}).

\paragraph{Subpool Selection (\Cref{alg:topk}).}
Finally, the $\subpoolsize = \min\{\size{\avgnnset}, \maxsubpoolsize\}$ pool instances with the highest similarity score are used as the subpool $\subpool$ to run the \AL strategy.
The hyper-parameter $\maxsubpoolsize$ controls the maximum size of the pool. In our experiments, we show that values as small as \q{1}{\thousand} are enough to achieve good performance. This step is responsible for the main speed-up provided by \name.

\vspace{.5em}
\begin{tcolorbox}[title={Tips: Hyper-parameters Settings}]
    The number of anchors per class $\numanchors$ and the number of neighbours retrieved per each class $\numneighbours$ should both be set to small values. Specifically, we suggest setting $\numanchors$ to \integer{5}-\integer{20} and $\numneighbours$ to \integer{20}-\integer{50}. The intuition is that as $\numanchors$ increases the subpool tends to be similar to the labelled set, thus failing to break the path dependence. As $\numneighbours$ increases the subpool tends to be dominated by majority instances due to the imbalance, thus reducing the chance of selecting minority instances.
\end{tcolorbox}

% ==================
% EXPERIMENTAL SETUP
% ==================
\vspace{.5em}
\section{Experimental Setup}\label{sec:experimental_setup}

\input{assets/tables/model_summary}

We mimic a realistic and interactive annotation setting \citep{grieshaber_fine-tuning_2020, yuan_cold-start_2020, maekawa_low-resource_2022, schroder_revisiting_2022} characterised by a budget of \q{5}{\thousand}\footnote{Throughout the paper, we use shorthands for units, that is, \q{}{\thousand} (thousands), \q{}{\million} (millions), and \q{}{\hour}(hours).} annotations and label \integer{25} instances per iteration\footnote{Note that this setup differs from previous \AL studies \citep[\emph{inter alia}]{margatina_active_2021,margatina_importance_2022, citovsky_batch_2021} that label a large portion of the pool per iteration (e.g., \q{1}{\percent}) and assume a large validation set throughout the \AL process.}.
We start with an initial set of \integer{100} instances $\dataset_0$, containing \integer{5} from each minority class and the rest from the majority, chosen randomly. We do not assume access to a validation set. In each iteration, we re-initialise the model and train all parameters. We limit each run (i.e., training, instance selection, and testing) to \q{6}{\hour} to allow for thorough experimentation while complying with our computational budget.\footnote{Datasets and experimental artefacts available at \href{https://huggingface.co/collections/pietrolesci/anchoral-66103ace42da659656c635d2}{\myemph{huggingface.co/collections/pietrolesci/anchoral}}.}

\input{assets/tables/small_table}

\paragraph{Evaluation.}
We use the (macro-averaged) F1-score on the minority class(es) as our predictive accuracy metric. For completeness, we report the majority class performance, too. We evaluate the model on a held-out set after each iteration and report the area under the learning curve (AUC) computed using the trapezoidal rule\footnote{As implemented in \href{https://numpy.org/doc/stable/reference/generated/numpy.trapz.html}{\myemph{numpy.trapz}}.}. Moreover, we report the total (across iterations) instance selection time as a proxy of the total annotators' waiting time.
%, assuming uniform instance annotation difficulty \citep{settles_active_2012}. 
To allow for a more robust comparison, we use \integer{2} random seeds for each major source of randomness: model initialisation, data ordering, and initial data selection resulting in \integer{8} runs per experiment and report the median and interquartile range \citep{liu_few-shot_2022}.\looseness=-1

\paragraph{Baselines.}
We compare \name with two pool filtering methods. \randomsubset \citep{ertekin_learning_2007} samples $\subpoolsize$ instances uniformly at random from the pool; we set $\subpoolsize\mathop{=}\q{10}{\thousand}$\footnote{As it would be done in a real setting, we searched over a grid of values in \q{5}{\thousand}-\q{20}{\thousand} for a small budget of \integer{200} labelled instances and selected the best (i.e., speed and AUC) $\subpoolsize$.}. \seals \citep{coleman_similarity_2022} limits the pool to the set of $\numneighbours\mathop{=}\integer{50}$ neighbours of the currently labelled data: after every instance is selected, its $\numneighbours$-nearest neighbours are added to the pool. Moreover, we compare against standard \AL (\noop) which considers the entire pool and random sampling (\random).

\paragraph{\AL Strategies.}
We use one \AL strategy from each type of \AL approach \citep{dasgupta_two_2011}. \entropy \citep[uncertainty-based]{joshi_multi-class_2009} selects instances with the highest predictive entropy. \ftbertkm \citep[diversity-based]{yuan_cold-start_2020} chooses instances nearest to cluster centres obtained running \kmeans clustering with $k$ equal to the number of instances to label. Finally, \badge \citep[hybrid]{ash_deep_2020} selects instances using the \kmeanspp initialisation scheme applied to the gradient embeddings. More details in \Cref{app:experimental_details}.

\paragraph{Models.}
We consider \integer{6} models of different sizes and architectures (\Cref{tab:model_summary}). Following the \myemph{transformers} library \citep{wolf_transformers_2020} implementation, we add a linear layer to the model representations: the \code{[CLS]} token for encoders, the last non-padding token for decoders, and the end-of-sequence token for encoder-decoders. Training details in \Cref{app:experimental_details}.

\paragraph{Datasets.}
We consider \integer{4} text classification tasks, both binary and multiclass. Using the \Amazoncat dataset \citep{mcauley_hidden_2013} we construct the \amazonagri task, selecting the \enquote{agriculture} (\q{.09}{\percent}) category as target, and the \amazonmulti task which also has \enquote{archaeology} (\q{0.09}{\percent}), \enquote{audio} (\q{.56}{\percent}), and \enquote{philosophy} (\q{0.78}{\percent}). Moreover, we consider the \wikitoxic \citep{jigsaw} and \agnewsbus---i.e., \Agnews \citep{zhang_character-level_2015} binarised using the \enquote{business} topic as the target--- tasks. We make them imbalanced downsampling their minority class to \q{1}{\percent}. More details in \Cref{app:data}.

% =======
% RESULTS
% =======
\section{Results}

\input{assets/images/minority_proportions}

\Cref{tab:bert_results} summarises a subset of our results for the \bertbase model and the \entropy strategy. The full results table in \Cref{app:results} provides similar insights for the other models and \AL strategies.
Since we limit the duration of each run to \q{6}{\hour}, slow pool filtering strategies might not be able to execute all iterations within the time budget. Thus, we report metrics at the end of the \q{6}{\hour} time budget (\enquote{Overall}) and at the biggest common iteration completed by all methods (\enquote{Budget-Matched}).\footnote{For example, if \seals only completes $\hat{t}$ steps within the available time while \name manages to complete all iterations, the Budget-Matched performance compares the results after $\hat{t}\mathop{\times}\querysize$ instances are annotated.}
For each experiment, we report the total annotated budget within the \q{6}{\hour} limit (\enquote{Budget}), the AUC of the (macro-averaged) F1-score on the majority (\enquote{Majority}) and minority (\enquote{Minority}) classes, and the total instances selection time (\enquote{Time}).

\paragraph{Cost Efficiency.}
Broadly, the overall annotation cost can be divided into the cost of the compute needed to run inference on the pool and the cost of the human annotations.
Pool filtering methods reduce both cost components by considering only a subset of the pool in each iteration: fewer instances require less compute and are faster to process which makes annotators spend less time waiting for instances to label resulting in more annotations within the same time budget.
\name is the fastest method overall, reducing the total annotators' waiting time (\enquote{Time} column) from hours to less than \integer{5} minutes and labelling the most instances within the time budget (\enquote{Budget} column).

\name is faster because it returns a smaller subpool than the other methods (\Cref{fig:glance} panel a). For example, \name returns \q{<1}{\thousand} instances at each iteration for the binary tasks while in \seals, by design, the subpool grows across iterations and for \randomsubset a larger subpool is required to achieve good performance.
In theory, \randomsubset can be made arbitrarily fast by choosing a small $\subpoolsize$ but it comes with high performance costs. For example, setting $\subpoolsize = \q{1}{\thousand}$, similar to \name, results in a dramatic performance drop (\cref{tab:ablations} last row). In the limiting case of setting $\subpoolsize = \querysize$ \randomsubset reduces to \random.

\paragraph{Performance.}
\name is the best-performing method for both minority and majority classes per time- (\enquote{Overall} columns) and budget-unit (\enquote{Budget-Matched} columns) across models and \AL strategies.
The only exception is \agnewsbus where it slightly underperforms \seals on the minority class(es). We hypothesise that this might be an artefact of the downsampling procedure we applied to this task to make it imbalanced: minority instances may be near the initial decision boundary, making exploration of the input space detrimental.
Furthermore, we notice that \randomsubset is a stronger baseline than reported by \citet{coleman_similarity_2022}. We hypothesise that this discrepancy depends on the fact that we train the entire model rather than only the classification head.

\input{assets/images/subpool_minority_proportions}

\paragraph{Discovery of Minority Instances.}
Although our primary goal in applying \AL is to induce a good predictive model, we also report the number of minority instances discovered by each method in \Cref{fig:minority_proportions}, which is likely to be indicative of the usefulness of the annotations when re-used for training other models. At every iteration, \name discovers more minority instances than the baselines resulting in a more balanced labelled set.

\begin{tcolorbox}[title={Contributions}]
    \name is the fastest method, reducing the total selection time from hours to minutes, thus allowing for an interactive annotation setup. It is (often) the best-performing, reaching higher performance in less time and with fewer annotations. Finally, it discovers the most minority instances resulting in more balanced labelled sets.
\end{tcolorbox}

% ========
% ANALYSIS
% ========
\section{Analysis}\label{sec:analysis}

\input{assets/tables/ablation_table}

In this section, we study \name's anchoring mechanism. This mechanism is the core component allowing \name to reach higher performance despite using a fixed-sized, small subpool. Everything else being equal, if the anchoring mechanism is turned off the performance is dramatically affected, as shown in the last row of \Cref{tab:ablations} which reports the performance of \name with no anchoring. We analyse the composition of the subpool at each iteration and test the effects of different hyper-parameters settings on performance by experimenting with the \bertbase model on the \amazonagri dataset using the \entropy \AL strategy.

\subsection{Subpool Composition}

The anchoring mechanism determines the composition of the subpool. In \Cref{fig:subpool_minority_proportions} we report the proportion of minority instances in the subpool returned by the methods considered.
We observe that, across iterations, \name consistently returns subpools with more minority instances; this results in more balanced labelled sets, as shown in \Cref{fig:minority_proportions}, and ultimately better performance. Therefore, the key intuition in \name is to return a more balanced subpool which allows any \AL strategy to discover and select minority instances more easily, without the need for a large subpool (e.g., unlike \randomsubset) which reduces the instance selection time.

\subsection{Ablations}

The anchoring mechanism is controlled by three hyperparameters: the number of anchors selected per class ($\numanchors$), the anchor selection strategy ($\anchorstrategy$), and the number of neighbours retrieved from the subpool per anchor ($\numneighbours$). Together, $\numanchors$ and $\numneighbours$ control the size of the resulting subpool while $\anchorstrategy$ determines which anchors are selected and, thus, which part of the input space to explore. In \Cref{sec:methodology} we presented our default settings; here we motivate their choice.

\paragraph{Anchor Selection Strategy ($\anchorstrategy$).}
We experiment with different anchor selection strategies for both majority ($\anchorstrategy_{\mathtt{maj}}$) and minority ($\anchorstrategy_{\mathtt{min}}$) anchors (\Cref{tab:ablations} row 2-3). As the alternative strategy, we use the entropy of the model. The reasoning is as follows: since the model knows the majority better than the minority class(es) it might be informative in choosing majority instances near the decision boundary. Overall, we report a negative effect, even more pronounced when entropy is used to select minority anchors too. We hypothesise that using a model-agnostic anchors selection strategy (e.g., \kmeanspp) avoids propagating the initial biases of the model in the selection of the instances.

\paragraph{Number of Anchors ($\numanchors$).}
We vary the number of anchors from \integer{10} to \integer{50} and \integer{100} (\Cref{tab:ablations} row 4-5). First, note that using all the labelled data as anchors at each iteration becomes quickly impractical (e.g., like \seals); since only a few iterations are completed within the \q{6}{\hour} limit, the performance is almost zero and we omit it. Second, there is a negative correlation between the number of anchors and performance. We hypothesise this is due to the bigger and more imbalanced resulting subpool which decreases the benefits of \name.

\paragraph{Number of Neighbours ($\numneighbours$).}
We change the default number of neighbours from \integer{50} to \integer{500} and \q{5}{\thousand} (\Cref{tab:ablations} row 6-7). Performance degrades as we retrieve more neighbours from the pool. Moreover, as the resulting subpool is bigger, instance selection time increases even though it plateaus after \integer{500} as the anchors retrieve the same instances that are then aggregated (\Cref{alg:algorithm} \Cref{alg:avg}).

% ===========
% CONCLUSIONS
% ===========
\section{Conclusions}

We propose \name, a novel pool filtering method designed to scale \AL to large pools while addressing class imbalance. \name uses the semantic representation capabilities of language models to explore the input space and create a fixed-sized, smaller, more balanced, and different subpool in each iteration. By running the \AL strategy on the subpool, \name promotes the discovery of minority instances, prevents overfitting the initial labelled set, and obtains a constant instance selection time, independently of the original pool size.

% ===========
% LIMITATIONS
% ===========
\clearpage
\section*{Limitations}

In this section, we discuss the limitations of the scope of this work and the threats to the external validity of our results.

\paragraph{Languages.}
We experimented with a limited number of languages (only English). We do not have experimental evidence that our method can work for other languages for which good embedding models are not available. Still, our approach has been built without any language-specific constraints or resources. Our method can be applied to any other language for which these resources are available.\looseness=-1

\paragraph{Realism.}
Recent \AL research emphasises the empirical evaluation of classifier performance resulting from simulated experiments. However, this idealised setting tacitly makes assumptions that cannot be true in real-world settings \citep{margatina-aletras-2023-limitations}; for example, a perfect oracle, uniform annotation difficulty, and the possibility to monitor the performance of the \AL strategy on a test set while training \citep{wallace_active_2010, levonian_trade-offs_2022}. Our paper suffers from these limitations too, even though we strived to address the annotators' waiting time issue. We leave for future work exploring methods to make \name more suited for practical use in real-world annotation settings.

% ================
% ACKNOWLEDGEMENTS
% ================
\section*{Acknowledgements}
\setlength{\intextsep}{0pt}
\setlength{\columnsep}{8pt}
\begin{wrapfigure}{l}{0.44\columnwidth}
    %\frame{
    \includegraphics[width=0.44\columnwidth]{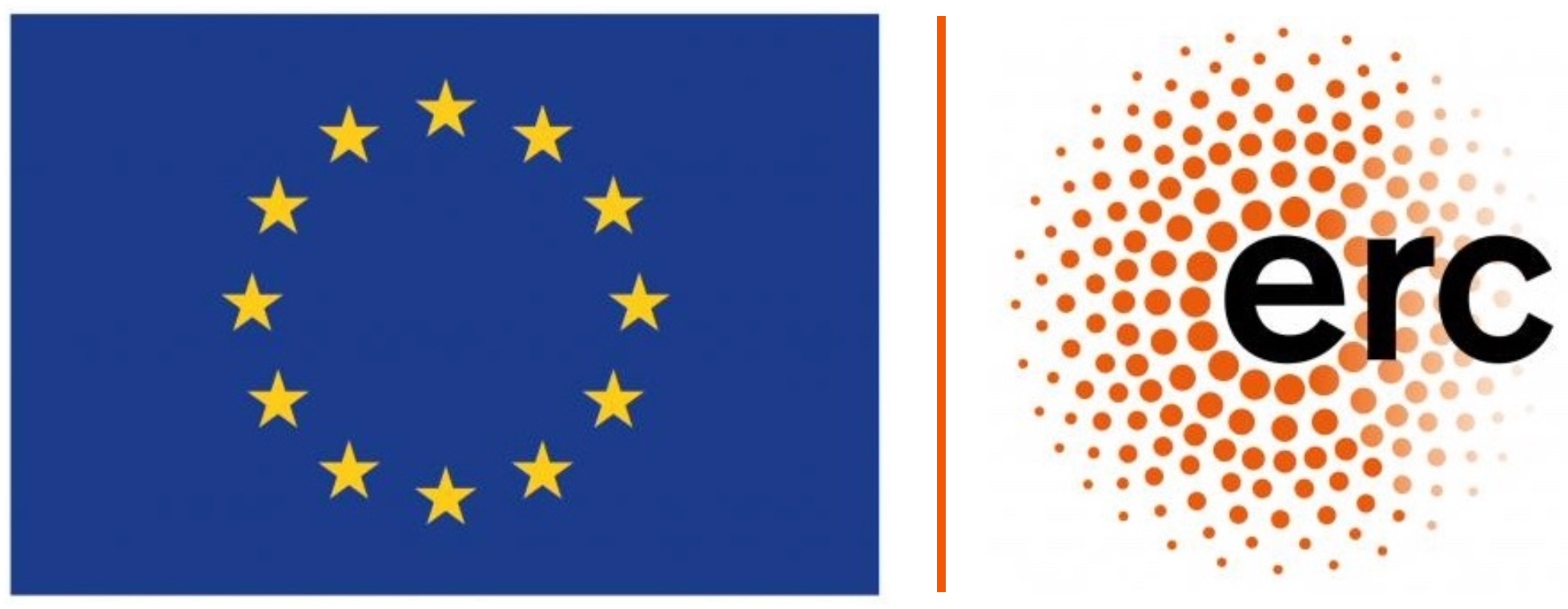}
    %}
\end{wrapfigure}
This project has received funding from the European Research Council (ERC) under the European Union’s Horizon 2020 Research and Innovation programme grant AVeriTeC (Grant agreement No. 865958).
We thank the anonymous reviewers for their helpful
questions and comments that helped us improve the paper. We thank Tiago Pimentel, Davide Lesci, and Marco Lesci for their help in proofreading the final version of the paper.

%% file: assets/images/method.tex
\begin{figure}
    \centering
    \includegraphics[width=\columnwidth]{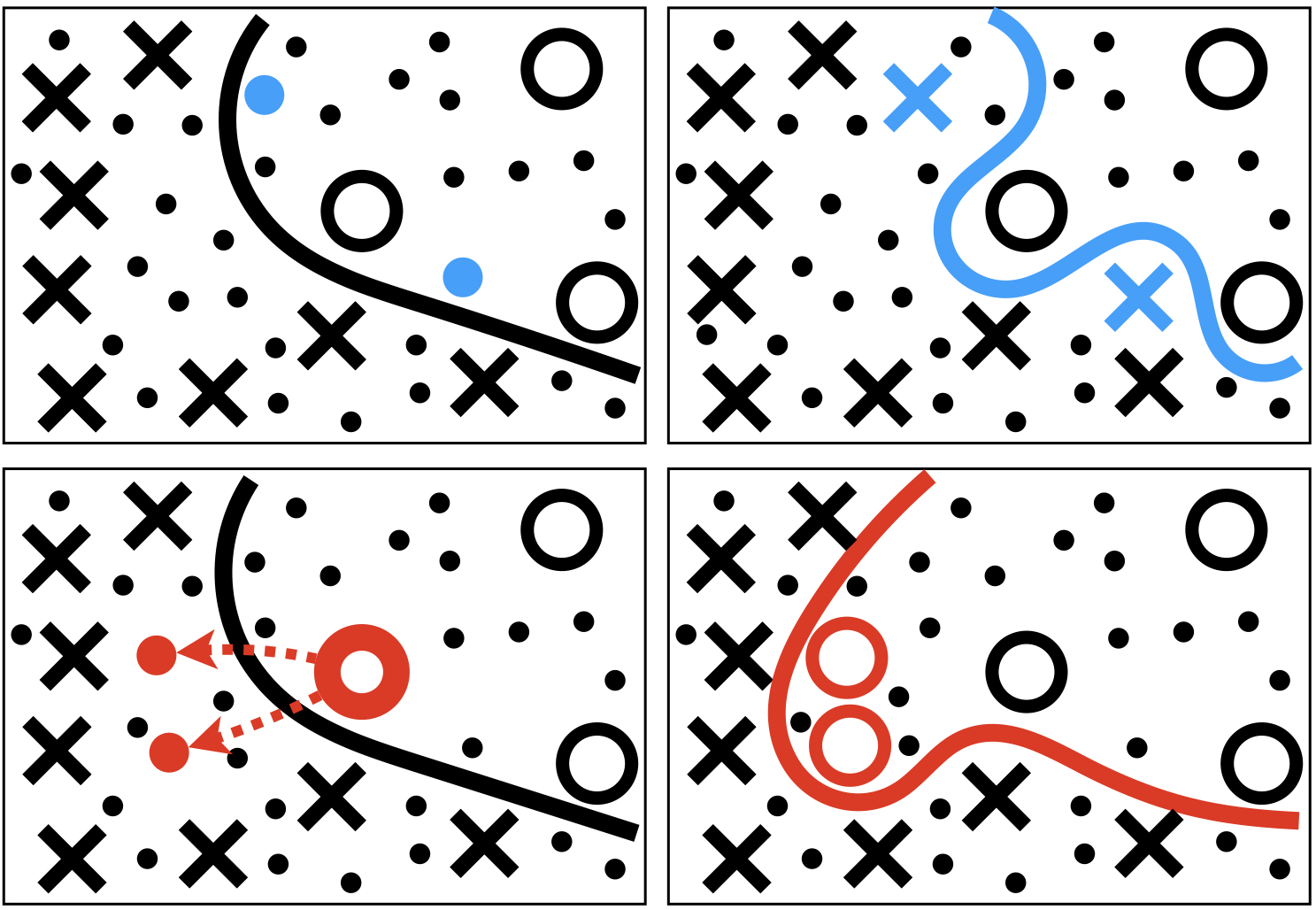}
    \caption{%
    \name (intuition). Binary classification task where $\cerchio$, $\cross$, and $\point$ are labelled minority, labelled majority, and unlabelled instances. The black (left) and coloured (right) lines denote the initial and final decision boundary. Typical \AL (top) selects instances near the current boundary.
    \name (bottom) anchors the selection to labelled instance(s) (bold red $\cerchio$) and discovers a new minority cluster.}
    \label{fig:method}
\end{figure}

%% file: assets/images/initial_plot.tex
\begin{figure*}[!ht]
    \centering
    \includegraphics[width=\textwidth]{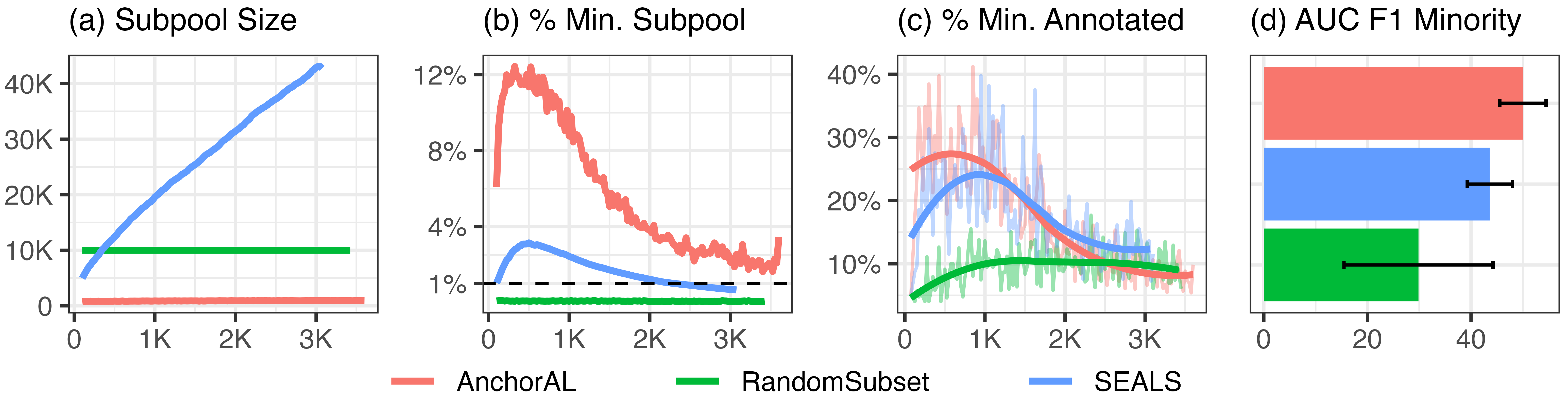}
    \caption{\name (this paper) vs \randomsubset~\citep{ertekin_learning_2007} and \seals~\citep{coleman_similarity_2022} on an imbalanced (\q{<1}{\percent}) binary classification task (\amazonagri) for the \albertbase model and the \entropy strategy. \name keeps the subpool size fixed and small across iterations (x-axis) (a) which reduces computational and annotation costs (i.e., annotators' waiting time). The subpool is more balanced (b) and allows the \AL strategy to discover minority instances earlier (c) and reach better performance (d).}
    \label{fig:glance}
    \vspace*{-.2em}
\end{figure*}

%% file: assets/alg_anchoral.tex
\begin{algorithm*}[!t]

    \centering

    \caption{\textbf{\name}}

    \label{alg:algorithm}

    \fontsize{10}{12pt}
    \begin{algorithmic}[1]
        \Require \AL strategy $\strategy$, anchor selection strategy $\anchorstrategy$, per-class number of anchors $\numanchors$,
        per-anchor number of neighbours $\numneighbours$, and maximum subpool size $\maxsubpoolsize$.

        \Ensure Unlabelled pool $\pool_0$, initial labelled set $\dataset_0$, oracle $\oracle$, dense index $\myindex$.% and nearest neighbours search implementation $\knn$.

        \smallskip

        \For{$t$ to $\numrounds$}

        \State $\model_t = \model.\trainfn(\dataset_{t})$
        \InlineComment{Train model on the available data}

        \State $\anchorset_t = \emptyset$ \label{alg:anchorselect_start}

        \For{$\class$ in $\{1,...,\numclasses\}$}

        \State $\anchorset_t \leftarrow \anchorstrategy\big(\{\vx \mid y \mathop{=} \class \;\; \forall \; (\vx, y) \in \dataset_t\}; \numanchors  \big)$
        \InlineComment{Select $\numanchors$ anchors per class}

        \EndFor \label{alg:anchorselect_end}

        \State $\nnset_t = \myindex.\knn(\anchorset_t, \pool_t; \numneighbours)$ \label{alg:knn}
        \InlineComment{Retrieve $\numneighbours$ neighbours per anchor from the pool}

        \State $\avgnnset_t = \aggfn(\nnset_t)$ \label{alg:avg}
        \InlineComment{Average similarity scores per each pool instance}

        \State $\subpool_t = \topk(\avgnnset_t; \subpoolsize)$ where $\subpoolsize = \min\{\size{\avgnnset}, \maxsubpoolsize\}$
        \InlineComment{Keep the top-$\subpoolsize$ instances} \label{alg:topk}

        \State $\queryset_t = \strategy(\model_t, \subpool_t; \querysize)$ where $\querysize = \lfloor\budget / \numrounds\rfloor$
        \InlineComment{Select $\querysize$ instances to label from the subpool}

        \State $\dataset_{t+1} = \dataset_t \cup \{(\poolx, \oracle(\poolx)) \mid \poolx \in \queryset_t\}$
        \InlineComment{Label queried instances and add to training set}

        \State $\pool_{t+1} = \pool_t \setminus \queryset_t$
        \InlineComment{Remove labelled instances from the pool}

        \EndFor
    \end{algorithmic}

\end{algorithm*}

%% file: assets/tables/model_summary.tex
\begin{table}[!t]
  \centering
  \small

  \begin{adjustbox}{width=\columnwidth}

    \begin{tabular}{llr}
      \toprule
      \textbf{Type}               & \textbf{Model}                                                                                         & \textbf{\# Params} \\
      \midrule
      \multirow[t]{6}{*}{Encoder} & \href{https://huggingface.co/bert-base-uncased}{\bertbase}~\citep{devlin_bert_2019}                    & \q{110}{\million}  \\
                                  & \href{https://huggingface.co/google/bert_uncased_L-4_H-256_A-4}{\berttiny}~\citep{turc_well-read_2019} & \q{4.4}{\million}  \\
                                  & \href{https://huggingface.co/albert-base-v2}{\albertbase} ~\citep{lan_albert_2019}                     & \q{12}{\million}   \\
                                  & \href{https://huggingface.co/microsoft/deberta-v3-base}{\debertabase} ~\citep{he_debertav3_2023}       & \q{86}{\million}   \\
      Decoder                     & \href{https://huggingface.co/gpt2}{\gpt} ~\citep{radford_2019_language}                                & \q{117}{\million}  \\
      Enc-Dec                     & \href{https://huggingface.co/t5-base}{\tf} ~\citep{raffel_exploring_2020}                              & \q{220}{\million}  \\
      \bottomrule
    \end{tabular}

  \end{adjustbox}
  \caption{Models overview. We use the checkpoints available on the HuggingFace Hub and loaded using the \href{https://huggingface.co/docs/transformers/v4.36.0/en/model_doc/auto}{\myemph{AutoModelForSequenceClassification}} class.}
  \label{tab:model_summary}
\end{table}

%% file: assets/tables/small_table.tex
\begin{table*}[!ht]
    \centering
    \small

    \begin{adjustbox}{width=\textwidth}
        \begin{tabular}{llrrrrrr}

            \toprule

                                             &                         & \multicolumn{4}{c}{\textbf{Overall}} & \multicolumn{2}{c}{\textbf{Budget-Matched}}                                                                                                                                                                                \\

            \cmidrule(lr){3-6}\cmidrule(lr){7-8}

            \textbf{Dataset}                 & \textbf{Pool Filtering} & \textbf{Budget}                      & \textbf{Majority}                           & \textbf{Minority}                        & \textbf{Time} ($\downarrow$)            & \textbf{Majority}                            & \textbf{Minority}                        \\

            \midrule

            \multirow[t]{4}{*}{\amazonagri}  & \name                   & \q{3.7}{\thousand}                   & $\float[1]{142.42}_{\pm\float[1]{0.04}}$    & $\float[1]{78.05}_{\pm\float[1]{1.0}}$   & $\float[1]{4.4}_{\pm\float[1]{0.05}}$   & $\float[1]{119.79}_{\pm\float[1]{0.05}}$     & $\float[1]{64.87}_{\pm\float[1]{0.89}}$  \\
                                             & \randomsubset           & \q{3.6}{\thousand}                   & $\float[1]{138.26}_{\pm\float[1]{0.06}}$    & $\float[1]{64.96}_{\pm\float[1]{3.18}}$  & $\float[1]{28.96}_{\pm\float[1]{0.19}}$ & $\float[1]{119.59}_{\pm\float[1]{0.05}}$     & $\float[1]{54.26}_{\pm\float[1]{1.95}}$  \\
                                             & \seals                  & \q{3.2}{\thousand}                   & $\float[1]{123.72}_{\pm\float[1]{0.02}}$    & $\float[1]{65.59}_{\pm\float[1]{0.78}}$  & $\float[1]{85.37}_{\pm\float[1]{1.28}}$ & $\float[1]{119.79}_{\pm\float[1]{0.03}}$     & $\float[1]{63.67}_{\pm\float[1]{1.44}}$  \\
                                             & \noop                   & \integer{475}                        & $\float[1]{14.62}_{\pm\float[1]{0.0}}$      & $\float[1]{2.9}_{\pm\float[1]{0.0}}$     & \q{6}{\hour}                            & $\float[1]{14.62}_{\pm\float[1]{0.0}}$       & $\float[1]{2.9}_{\pm\float[1]{0.0}}$     \\

            \midrule

            \multirow[t]{4}{*}{\amazonmulti} & \name                   & \q{3.6}{\thousand}                   & $\float[1]{109.24}_{\pm\float[1]{1.69}}$    & $\float[1]{77.85}_{\pm\float[1]{9.94}}$  & $\float[1]{7.78}_{\pm\float[1]{0.22}}$  & $\float[1]{92.66}_{\pm\float[1]{1.72}}$      & $\float[1]{65.27}_{\pm\float[1]{12.36}}$ \\
                                             & \randomsubset           & \q{3.5}{\thousand}                   & $\float[1]{106.17}_{\pm\float[1]{0.62}}$    & $\float[1]{74.73}_{\pm\float[1]{14.18}}$ & $\float[1]{26.63}_{\pm\float[1]{0.28}}$ & $\float[1]{92.78}_{\pm\float[1]{0.64}}$      & $\float[1]{64.99}_{\pm\float[1]{12.82}}$ \\
                                             & \seals                  & \q{3.2}{\thousand}                   & $\float[1]{95.33}_{\pm\float[1]{1.58}}$     & $\float[1]{69.2}_{\pm\float[1]{14.71}}$  & $\float[1]{83.82}_{\pm\float[1]{2.67}}$ & $\float[1]{93.18}_{\pm\float[1]{1.56}}$      & $\float[1]{67.84}_{\pm\float[1]{12.61}}$ \\
                                             & \noop                   & \integer{475}                        & $\float[1]{10.84}_{\pm\float[1]{0.29}}$     & $\float[1]{4.7}_{\pm\float[1]{3.24}}$    & \q{6}{\hour}                            & $\float[1]{10.84}_{\pm\float[1]{0.29}}$      & $\float[1]{4.7}_{\pm\float[1]{3.24}}$    \\

            \midrule

            \multirow[t]{4}{*}{\wikitoxic}   & \name                   & \q{4.2}{\thousand}                   & $\float[1]{128.07}_{\pm\float[1]{0.55}}$    & $\float[1]{119.0}_{\pm\float[1]{1.17}}$  & $\float[1]{2.55}_{\pm\float[1]{0.0}}$   & $\float[1]{106.62}_{\pm\float[1]{0.45}}$     & $\float[1]{99.03}_{\pm\float[1]{1.09}}$  \\
                                             & \randomsubset           & \q{4.0}{\thousand}                   & $\float[1]{119.36}_{\pm\float[1]{0.24}}$    & $\float[1]{107.36}_{\pm\float[1]{0.75}}$ & $\float[1]{20.39}_{\pm\float[1]{0.84}}$ & $\float[1]{104.67}_{\pm\float[1]{0.39}}$     & $\float[1]{94.56}_{\pm\float[1]{1.38}}$  \\
                                             & \seals                  & \q{3.6}{\thousand}                   & $\float[1]{109.59}_{\pm\float[1]{0.71}}$    & $\float[1]{100.58}_{\pm\float[1]{1.84}}$ & $\float[1]{60.89}_{\pm\float[1]{1.54}}$ & $\float[1]{105.88}_{\pm\float[1]{0.46}}$     & $\float[1]{97.59}_{\pm\float[1]{1.27}}$  \\
                                             & \noop                   & \q{2.9}{\thousand}                   & $\float[1]{88.28}_{\pm\float[1]{0.39}}$     & $\float[1]{80.82}_{\pm\float[1]{0.96}}$  & \q{6}{\hour}                            & $\float[1]{88.28}_{\pm\float[1]{0.39}}$      & $\float[1]{80.82}_{\pm\float[1]{0.96}}$  \\

            \midrule

            \multirow[t]{3}{*}{\agnewsbus}   & \name                   & \q{5}{\thousand}                     & $\float[1]{179.18}_{\pm\float[1]{0.12}}$    & $\float[1]{118.2}_{\pm\float[1]{1.26}}$  & $\float[1]{2.66}_{\pm\float[1]{0.1}}$   &                                              &                                          \\
                                             & \randomsubset           & \q{5}{\thousand}                     & $\float[1]{179.21}_{\pm\float[1]{0.39}}$    & $\float[1]{117.99}_{\pm\float[1]{4.03}}$ & $\float[1]{21.53}_{\pm\float[1]{0.1}}$  & \multicolumn{2}{c}{\textit{same as Overall}}                                            \\
                                             & \seals                  & \q{5}{\thousand}                     & $\float[1]{179.55}_{\pm\float[1]{0.06}}$    & $\float[1]{120.8}_{\pm\float[1]{0.79}}$  & $\float[1]{62.16}_{\pm\float[1]{1.25}}$ &                                              &                                          \\

            \bottomrule
        \end{tabular}
    \end{adjustbox}

    \caption{AUC of the F1-score, total annotated budget, and total instance selection time for the \bertbase model and \entropy \AL strategy. Median and interquartile range across \integer{8} runs. %Results at the end of the \q{6}{\hour} budget for each run (Overall) and at the biggest common annotated budget by all pool filtering strategies (Budget-Matched).
    }
    \label{tab:bert_results}
\end{table*}

%% file: assets/images/minority_proportions.tex
\begin{figure}
    \centering
    \includegraphics[width=\columnwidth]{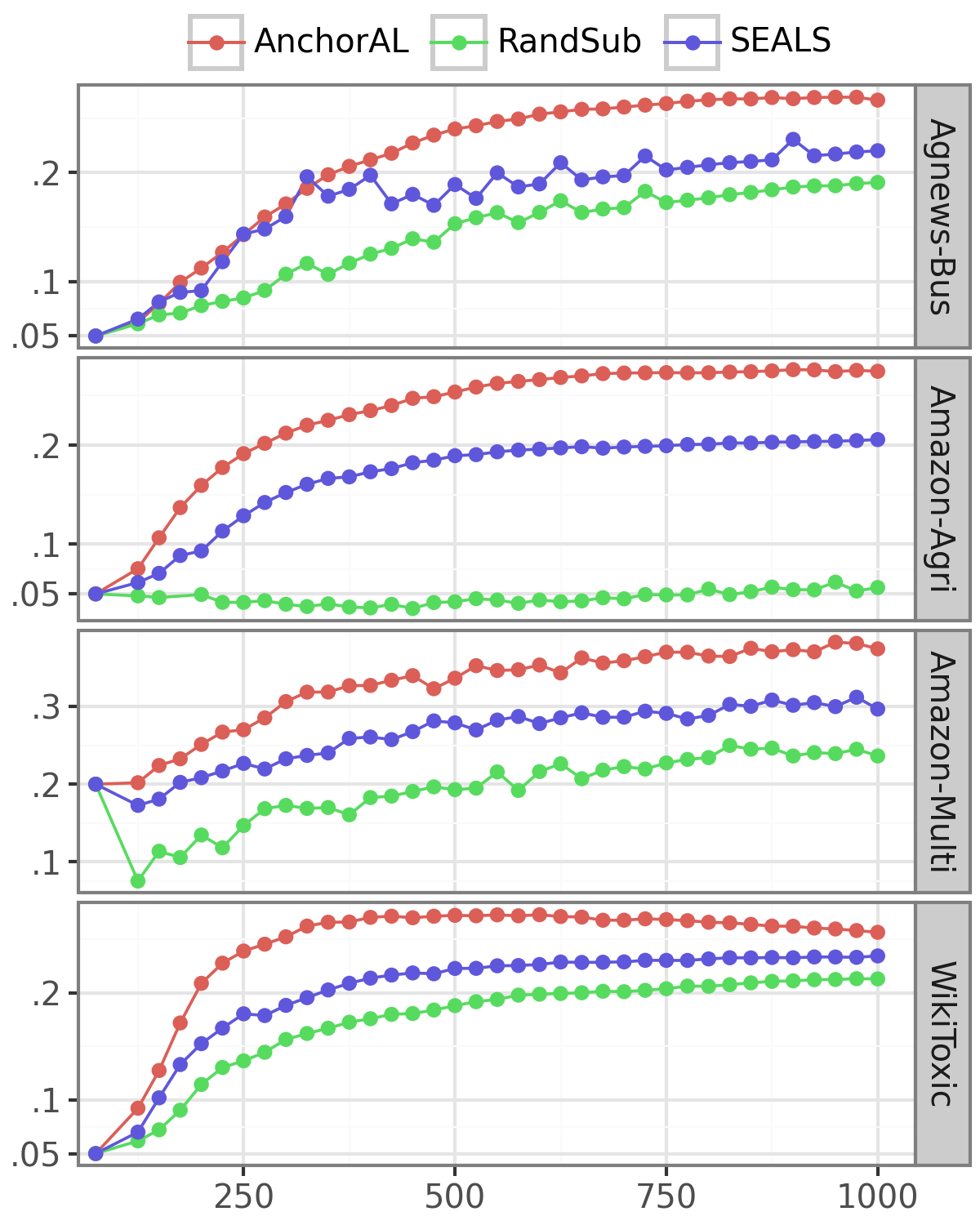}
    \caption{Proportion of minority instances (y-axis) discovered within \q{1}{\thousand} annotations (x-axis). Geometric average across all \AL strategies considered for \bertbase.}
    \label{fig:minority_proportions}
\end{figure}

%% file: assets/images/subpool_minority_proportions.tex
\begin{figure}
    \centering
    \includegraphics[width=\columnwidth]{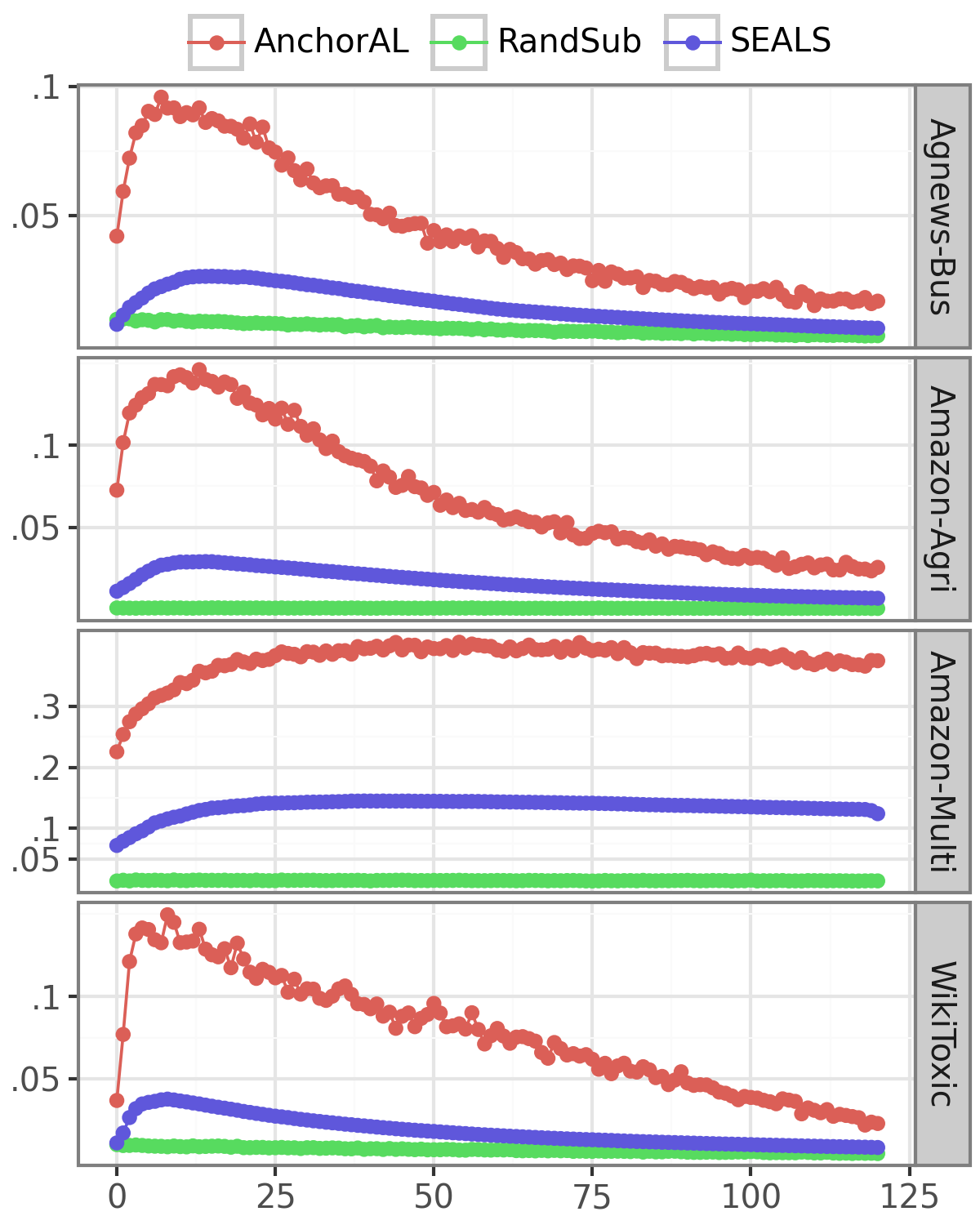}
    \caption{Proportion of minority instances (y-axis) in the subpool at each iteration (x-axis). Geometric average across all \AL strategies considered for \bertbase.}
    \label{fig:subpool_minority_proportions}
\end{figure}

%% file: assets/tables/ablation_table.tex
\begin{table}[!t]
    \centering
    %\fontsize{8.5pt}{10pt}
    \small

    \begin{adjustbox}{width=\columnwidth}
        \begin{tabular}{llllrrr}

            \toprule

            $\anchorstrategy_{\mathtt{maj}}$ & $\anchorstrategy_{\mathtt{min}}$ & $\numanchors$                     & $\numneighbours$                     & \textbf{Maj.}                 & \textbf{Min.}                & \textbf{Time}                 \\

            \midrule

            \myemph{kM++}                    & \myemph{kM++}                    & \integer{10}                      & \integer{50}                         & \float[1]{134.53658667206764} & \float[1]{73.52097980305552} & \float[1]{4.148966918389003}  \\

            \midrule

            \myemph{Ent} \cellcolor{gray!30} & \myemph{kM++}                    & \integer{10}                      & \integer{50}                         & \float[1]{134.51048186421394} & \float[1]{71.54710494577885} & \float[1]{12.688079781333606} \\

            \myemph{Ent} \cellcolor{gray!30} & \myemph{Ent} \cellcolor{gray!30} & \integer{10}                      & \integer{50}                         & \float[1]{134.51048186421394} & \float[1]{71.44710494577885} & \float[1]{12.813388372461}    \\

            \midrule

            \myemph{kM++}                    & \myemph{kM++}                    & \integer{50} \cellcolor{gray!30}  & \integer{50}                         & \float[1]{134.54071651399136} & \float[1]{72.62231518840417} & \float[1]{14.734867099920908} \\

            \myemph{kM++}                    & \myemph{kM++}                    & \integer{100} \cellcolor{gray!30} & \integer{50}                         & \float[1]{134.53013283014297} & \float[1]{72.21376596391201} & \float[1]{24.825135803222658} \\

            \midrule

            \myemph{kM++}                    & \myemph{kM++}                    & \integer{10}                      & \integer{500}\cellcolor{gray!30}     & \float[1]{134.49540776014328} & \float[1]{70.65079025924206} & \float[1]{30.131086428960163} \\

            \myemph{kM++}                    & \myemph{kM++}                    & \integer{10}                      & \q{5}{\thousand} \cellcolor{gray!30} & \float[1]{134.45990166068077} & \float[1]{68.29706193879247} & \float[1]{39.04649945894877}  \\

            \midrule

            \multicolumn{4}{l}{No anchoring} & \float[1]{133.73706080019474}    & \float[1]{33.55195029289462}      & \float[1]{3.385290997227033}                                                                                                        \\

            \bottomrule
        \end{tabular}
    \end{adjustbox}

    \caption{Effect of hyper-parameters on \name's performance (F1-score) for the \bertbase model on the \amazonagri task using the \entropy \AL strategy. Defaults in the top row, changes highlighted in \colorbox{gray!30}{grey}.}
    \label{tab:ablations}
\end{table}

%% file: appendix.tex
\section{Experimental Details}\label{app:experimental_details}

In addition to the information already present in \Cref{sec:experimental_setup}, here we add more specific implementation details.

\subsection{Index Construction}\label{app:index_construction}

We embed each document using the MPNet encoder \citep{song_mpnet_2020}\footnote{\href{https://huggingface.co/sentence-transformers/all-mpnet-base-v2}{\myemph{huggingface.co/sentence-transformers/all-mpnet-base-v2}}.} available on the HuggingFace Hub and implemented through the Sentence-Transformers\footnote{\href{https://www.sbert.net/}{\myemph{sbert.net}}.} library \citep{reimers_sentence-bert_2019}.
Embedding a batch of \integer{1024} documents on our hardware (\Cref{app:hardware_details}) takes circa \float[1]{5.5} seconds.
For efficient dense retrieval, we use Hierarchical Navigable Small World (HNSW) approximate nearest neighbour search algorithm proposed by \citet{malkov_efficient_2018} and available through the \myemph{hnswlib}\footnote{\href{https://github.com/nmslib/hnswlib}{\myemph{github.com/nmslib/hnswlib}}.} library. When building the index we use cosine similarity as the distance metric and the following standard settings: $\mathtt{ef\_construction} = \integer{200}$, $\mathtt{ef} = \integer{200}$, and $\mathtt{M} = \integer{64}$ to control the speed/accuracy trade-off during the index construction, query time/accuracy trade-off, and the maximum number of outgoing connections in the graph, respectively.

\subsection{Training Details}
We use the \href{https://pytorch.org/docs/stable/generated/torch.optim.AdamW.html}{AdamW} optimiser \citep{loshchilov_decoupled_2018} with the default hyperparameters---as implemented in Pytorch, that is $\beta_1\mathop{=}.9$, $\beta_2\mathop{=}.999$, $\mathtt{weight\_decay}\mathop{=}.01$, $\mathtt{eps}\mathop{=}\snum{e-8}$---and set the learning rate to \snum{4e-5} (\snum{2e-4} for \berttiny) with a constant schedule and a batch size of \integer{32}.
We truncate the input sequences to \integer{512} tokens. We do not assume access to a validation set and, instead, train for \integer{10} epochs selecting the best model based on the training F1-score on the minority class(es). Since the number of batches in the initial iterations is limited ($\integer{3} \mathop{\approx} \nicefrac{\integer{100}}{\integer{32}}$), to allow the model to converge we set the minimum number of optimisation steps to \integer{100} and use early stopping with a minimum delta of \snum{e-5}.

\subsection{Reproducibility}\label{app:reproducibility}
We implement all experiments using the PyTorch \citep{paszke_pytorch_2019} framework. We assign a different pseudo-random number generator to data shuffling, model initialisation, \AL strategy, and pool filtering strategy. In addition, we use CUDA deterministic operations and set the seeds for pseudo-random number generators in PyTorch, Numpy \citep{harris_array_2020}, the Python Random module \citep{vanrossum_python_2009}, and for each multi-processing worker.

\subsection{Active Learning Strategies}

In this section we provide an overview of the \AL strategy used in the paper.

\paragraph{Entropy.}
The \entropy strategy scores instances by computing the predictive entropy of the probability distribution assigned by the model, that is
\begin{align*}
    H(\vx) = -\sum_{\class=1}^C \model(\vx)_\class\,\log\model(\vx)_\class
\end{align*}

\paragraph{FT-BERTKM.}
The \ftbertkm strategy computes instance representations using the trained model. These representations correspond to the output of the penultimate layer of the model $h(\vx)$, which is the input to the classification layer. Once these representations are computed and $l2$-normalised for each unlabelled instance, it runs \kmeans setting $k$ equal to the number of instances to select in each iteration. Finally, the cluster centres (or the closest instances in the embedding space) are selected.

\paragraph{BADGE.}
The goal of \badge is to sample a diverse and uncertain batch of points for training neural networks. The algorithm transforms data into representations that encode model confidence and then clusters these transformed points. First, an instance $\vx$ is passed through the trained model to obtain its predicted label
\begin{equation*}
    \hat{y} = \argmax \model(\vx)
\end{equation*}
Next, a gradient embedding of the last layer of the model $g_{\vx}$ is obtained with respect to the loss computed using the predicted labels as the target
\begin{align*}
    g_{\vx} & = \nabla_\theta \mathcal{L}\left(\model_\theta(\vx\right), \hat{y})
\end{align*}
where $\mathcal{L}$ is the cross-entropy loss and $\theta$ are the parameters of the last layer of the model. The gradient embeddings can be computed in closed form as
\begin{align*}
    (g_{\vx})_\class = \left[\model_\theta(\vx)_\class - \mathbbm{1}(\hat{y} = \class)\right] \; h_\theta(\vx)
\end{align*}
where $h(\cdot)$ computes the model representation that feeds into the last layer. The gradient embedding is a multi-dimensional tensor since we are computing the gradient with respect to all possible classes $\class\in\{1, ..., C\}$. Thus, the $\class$-th block of $g_{\vx}$ is the hidden representation $h(\vx)$ scaled by the difference between model confidence score $\model(\vx)_{\class}$ and an indicator function that indicates whether the predicted class $\hat{y}$ is label $\class$.
Finally, BADGE chooses a batch to sample by applying \kmeanspp on these (flattened) gradient embeddings. These embeddings consist of model confidence scores and hidden representations, so they encode information about both uncertainty and the data distribution. By applying \kmeanspp on the gradient embeddings, the chosen examples differ in feature representation and predictive uncertainty.

\subsection{Hardware Details}\label{app:hardware_details}

We use a server with one NVIDIA A100 80GB PCIe, 32 CPUs, and 32 GB of RAM for all experiments. Below, we report a subset of the output of the \myemph{lscpu} command:

\begin{tcolorbox}[left=5pt,right=5pt,top=5pt,bottom=5pt]
    \small
    \begin{verbatim}
Architecture:        x86_64
CPU op-mode(s):      32-bit, 64-bit
Address sizes:       46 bits physical, 
                     48 bits virtual
Byte Order:          Little Endian
CPU(s):              32
On-line CPU(s) list: 0-31
Vendor ID:           GenuineIntel
Model name:          Intel(R) Xeon(R)
                     Silver 4210R CPU
                     @ 2.40GHz
CPU family:          6
Model:               85
Thread(s) per core:  1
Core(s) per socket:  1
Socket(s):           8
Stepping:            7
BogoMIPS:            4800.11
\end{verbatim}
\end{tcolorbox}

% ============
% DATA DETAILS
% ============
\section{Data}\label{app:data}

Our data pipeline is divided into two steps: sourcing and preparation.

\subsection{Data Sourcing}
We download the raw datasets from their original sources or other comparable faithful sources (e.g., HuggingFace Hub only when the dataset is created by the original authors or the HuggingFace team).

\paragraph{\Amazoncat.}
The \Amazoncat dataset was released by \citet{mcauley_hidden_2013} and is composed of product descriptions and reviews classified into \q{13}{\thousand} multi-label categories. The dataset is split into \q{1.2}{\million} train and \q{300}{\thousand} evaluation instances. It is commonly used as an extreme classification benchmark \citep{you_attentionxml_2019}\footnote{\href{http://manikvarma.org/downloads/XC/XMLRepository.html}{\myemph{manikvarma.org/downloads/XC/XMLRepository.html}}.} where the goal is to classify an item into its categories. We download the raw data from the original data source\footnote{\href{https://drive.google.com/u/0/uc?id=17rVRDarPwlMpb3l5zof9h34FlwbpTu4l}{\myemph{drive.google.com/17rVRDarPwlMpb3l5zof9h34FlwbpTu4l}}.}.

\paragraph{\Agnews.}
The AG dataset\footnote{\href{http://groups.di.unipi.it/~gulli/AG_corpus_of_news_articles.html}{\myemph{groups.di.unipi.it/~gulli/AG_corpus_of_news_articles.html}}.} consists of more than \q{1}{\million} news articles written in English. The articles have been collected from more than \q{2}{\thousand} news sources by ComeToMyHead, an academic news search engine running since July 2004, over a period of more than one year. The \Agnews topic classification dataset is a curated version proposed in \citet{zhang_character-level_2015} and is constructed by choosing the \integer{4} largest classes from the original corpus: World, Sports, Business, Sci/Tech. The dataset is split into \q{120}{\thousand} train and \q{7.6}{\thousand} evaluation instances. We use the version available on the HuggingFace Hub\footnote{\href{https://huggingface.co/datasets/ag_news}{\myemph{huggingface.co/datasets/ag\_news}}.}.

\paragraph{\Wikitoxic.}
The \Wikitoxic dataset is an updated version of the Kaggle Toxic Comment dataset used in the homonymous 2017/2018 challenge\footnote{\href{https://www.kaggle.com/competitions/jigsaw-toxic-comment-classification-challenge/overview}{\myemph{kaggle.com/competitions/jigsaw-toxic-comment-classification-challenge/overview}}.}. It contains comments in English collected from Wikipedia forums and classifies them into two categories, Toxic and Non-toxic. The dataset is split into \q{128}{\thousand} train, \q{32}{\thousand} validation, and \q{64}{\thousand} test instances. We download the data from the HuggingFace Hub\footnote{\href{https://huggingface.co/datasets/OxAISH-AL-LLM/wiki_toxic}{\myemph{huggingface.co/datasets/OxAISH-AL-LLM/wiki_toxic}}.}.

\input{assets/tables/data_stats}

\subsection{Data Preparation}
Data preparation refers to the process of converting a dataset into a task, that is preparing the data for training and evaluation which entails encoding the labels to integers and tokenising the documents. First, we only use the respective training and test sets for all the datasets since we do not assume access to a fixed validation set when training the model in each iteration. Second, we tokenise the documents using the model-specific tokeniser available in the \myemph{tokenizers}\footnote{\href{https://github.com/huggingface/tokenizers}{\myemph{github.com/huggingface/tokenizers}}.} library \citep{wolf_transformers_2020}.
Third, we encode the labels by mapping each class to an integer. Finally, we optionally downsample the minority class to increase the imbalance. Below we report the process of creating the classification tasks we use in our experiments from the original datasets. In \Cref{tab:data_stats}, we report statistics about the tasks used in our experiments.

Since \Amazoncat is a multilabel dataset, we choose some of the labels and create binary and multiclass classification tasks. We create one binary classification task, \amazonagri, by choosing labels referring to the \enquote{agricultural sciences} topic: 246. If any of the topic-specific labels appear in the label set of a document we assign the document to the minority class. We create the multiclass task, \amazonmulti, by using the \enquote{agriculture} label;  the \enquote{philosophy} label corresponding to the indices 413, 3999, 4044, 5881, 8900, 8901, 10990, 13319; the \enquote{archaeology} label with indices 538, 539, 8601; and the \enquote{audio} label with index 677.

For \Agnews we choose a label and treat it as the minority class and we downsample it in the training set to increase the imbalance. We choose the \enquote{Business} class and create the \agnewsbus task. We make the minority class account for only \percentage[0]{0.01} of the training set.

Finally, for \Wikitoxic, being already a binary classification task, we only downsample the \enquote{tox} class to only account for \percentage[0]{0.01} of the training set.

\section{Full Table of Results}\label{app:results}

The table follows in the next pages and reports the
AUC of the F1-score, total annotated budget, and total instance selection time for all models and \AL strategies considered. Median and interquartile range across \integer{8} runs. Results are reported at the end of the \q{6}{\hour} budget for each run (\enquote{Overall}) and at the biggest common annotated budget by all pool filtering strategies (\enquote{Budget-Matched}); differences in budget highlighted in \colorbox{gray!30}{grey}.

\clearpage
\onecolumn

\input{assets/tables/main_table}

%% file: assets/tables/data_stats.tex
% squeeze more horizontal space between columns
\setlength{\tabcolsep}{2.5pt}

\begin{table}[!t]
    \centering
    \small
    \begin{adjustbox}{width=\columnwidth}

        % === TABLE === %

        \begin{tabular}{llrr|rr}
            \toprule
                                             &                    & \multicolumn{2}{c}{\textbf{Test}} & \multicolumn{2}{c}{\textbf{Train}}                                                              \\
            \textbf{Dataset}                 & \textbf{Label}     & \textbf{\#}                       & \textbf{\%}                         & \textbf{\#}       & \textbf{\%}                           \\

            \midrule

            \multirow[c]{3}{*}{\amazonagri}  & \code{Negative}    & \integer{5000}                    & \percentage[2]{0.9460737937559129}  & \integer{1185188} & \percentage[2]{0.9991140065366254}    \\
                                             & \code{Positive}    & \integer{285}                     & \percentage[2]{0.05392620624408704} & \integer{1051}    & \percentage[2]{0.0008859934633745814} \\
            \cmidrule{3-6}
                                             &                    & \integer{5285}                    & \percentage[0]{1.0}                 & \integer{1186239} & \percentage[0]{1.0}                   \\

            \midrule

            \multirow[c]{6}{*}{\amazonmulti} & \code{agriculture} & \integer{285}                     & \percentage[2]{0.0292187820381382}  & \integer{1051}    & \percentage[2]{0.0008859934633745814} \\
                                             & \code{archaeology} & \integer{256}                     & \percentage[2]{0.02624564281320484} & \integer{1010}    & \percentage[2]{0.0008514304452981229} \\
                                             & \code{audio}       & \integer{1759}                    & \percentage[2]{0.18033627229854418} & \integer{6651}    & \percentage[2]{0.005606795932354272}  \\
                                             & \code{others}      & \integer{5000}                    & \percentage[2]{0.5126102111954071}  & \integer{1168266} & \percentage[2]{0.9848487530758978}    \\
                                             & \code{philosophy}  & \integer{2454}                    & \percentage[2]{0.2515890916547058}  & \integer{9261}    & \percentage[2]{0.007807027083075164}  \\
            \cmidrule{3-6}
                                             &                    & \integer{9754}                    & \percentage[0]{1.0}                 & \integer{1186239} & \percentage[0]{1.0}                   \\

            \midrule

            \multirow[c]{3}{*}{\agnewsbus}   & \code{Negative}    & \integer{5700}                    & \percentage[2]{0.75}                & \integer{90000}   & \percentage[2]{0.989990100098999}     \\
                                             & \code{Positive}    & \integer{1900}                    & \percentage[2]{0.25}                & \integer{910}     & \percentage[2]{0.01000989990100099}   \\
            \cmidrule{3-6}
                                             &                    & \integer{7600}                    & \percentage[0]{1.0}                 & \integer{90910}   & \percentage[0]{1.0}                   \\

            \midrule

            \multirow[c]{3}{*}{\wikitoxic}   & \code{non}         & \integer{5000}                    & \percentage[2]{0.4447211598327848}  & \integer{114722}  & \percentage[2]{0.9899983603869487}    \\
                                             & \code{tox}         & \integer{6243}                    & \percentage[2]{0.5552788401672152}  & \integer{1159}    & \percentage[2]{0.01000163961305132}   \\
            \cmidrule{3-6}
                                             &                    & \integer{11243}                   & \percentage[0]{1.0}                 & \integer{115881}  & \percentage[0]{1.0}                   \\

            \bottomrule
        \end{tabular}

        % ============= %

    \end{adjustbox}
    \caption{Data statistics of the task considered.}
    %\vspace*{-1.em}
    \label{tab:data_stats}
\end{table}

%% file: assets/tables/main_table.tex
\begin{landscape}
    \begingroup
    \footnotesize  % <--- size of fonts in the table

    \begin{xltabular}{\linewidth}{XXXXXXXXXXXX}

        \toprule
        &  &  &  & \multicolumn{4}{c}{\textbf{Overall}} & \multicolumn{4}{c}{\textbf{Budget-Matched}} \\

        \cmidrule(lr){5-8}\cmidrule(lr){9-12}

        \textbf{Dataset} & \textbf{Model} & \textbf{\AL Strategy} & \textbf{Pool Filtering} & \textbf{Budget} & \textbf{Majority} & \textbf{Minority} & \textbf{Time} & \textbf{Budget} & \textbf{Majority} & \textbf{Minority} & \textbf{Time} \\

        \midrule\midrule
        \multirow[t]{25}{*}{\agnewsbus} & \multirow[t]{3}{*}{\albertbase} & \multirow[t]{3}{*}{\entropy} & \name & \q{5.0}{\thousand} & $\float[1]{176.52}_{\pm\float[1]{0.69}}$ & $\float[1]{96.92}_{\pm\float[1]{9.97}}$ & $\float[1]{2.73}_{\pm\float[1]{0.08}}$ & \q{5.0}{\thousand} & $\float[1]{176.52}_{\pm\float[1]{0.69}}$ & $\float[1]{96.92}_{\pm\float[1]{9.97}}$ & $\float[1]{2.73}_{\pm\float[1]{0.08}}$ \\
        &  &  & \randomsubset & \q{5.0}{\thousand} & $\float[1]{176.74}_{\pm\float[1]{1.77}}$ & $\float[1]{98.27}_{\pm\float[1]{5.47}}$ & $\float[1]{22.43}_{\pm\float[1]{0.5}}$ & \q{5.0}{\thousand} & $\float[1]{176.74}_{\pm\float[1]{1.77}}$ & $\float[1]{98.27}_{\pm\float[1]{5.47}}$ & $\float[1]{22.43}_{\pm\float[1]{0.5}}$ \\
        &  &  & \seals & \q{5.0}{\thousand} & $\float[1]{176.34}_{\pm\float[1]{1.23}}$ & $\float[1]{98.25}_{\pm\float[1]{4.09}}$ & $\float[1]{66.48}_{\pm\float[1]{1.18}}$ & \q{5.0}{\thousand} & $\float[1]{176.34}_{\pm\float[1]{1.23}}$ & $\float[1]{98.25}_{\pm\float[1]{4.09}}$ & $\float[1]{66.48}_{\pm\float[1]{1.18}}$ \\
        \cmidrule{2-12}
        & \multirow[t]{10}{*}{\bertbase} & \multirow[t]{3}{*}{\badge} & \name & \q{5.0}{\thousand} & $\float[1]{179.1}_{\pm\float[1]{0.3}}$ & $\float[1]{116.9}_{\pm\float[1]{2.29}}$ & $\float[1]{2.85}_{\pm\float[1]{0.06}}$ & \q{5.0}{\thousand} & $\float[1]{179.1}_{\pm\float[1]{0.3}}$ & $\float[1]{116.9}_{\pm\float[1]{2.29}}$ & $\float[1]{2.85}_{\pm\float[1]{0.06}}$ \\
        &  &  & \randomsubset & \q{5.0}{\thousand} & $\float[1]{179.32}_{\pm\float[1]{0.41}}$ & $\float[1]{118.76}_{\pm\float[1]{2.78}}$ & $\float[1]{23.7}_{\pm\float[1]{0.08}}$ & \q{5.0}{\thousand} & $\float[1]{179.32}_{\pm\float[1]{0.41}}$ & $\float[1]{118.76}_{\pm\float[1]{2.78}}$ & $\float[1]{23.7}_{\pm\float[1]{0.08}}$ \\
        &  &  & \seals & \q{5.0}{\thousand} & $\float[1]{179.57}_{\pm\float[1]{0.34}}$ & $\float[1]{121.09}_{\pm\float[1]{2.91}}$ & $\float[1]{74.0}_{\pm\float[1]{1.36}}$ & \q{5.0}{\thousand} & $\float[1]{179.57}_{\pm\float[1]{0.34}}$ & $\float[1]{121.09}_{\pm\float[1]{2.91}}$ & $\float[1]{74.0}_{\pm\float[1]{1.36}}$ \\
        \cmidrule{3-12}
        &  & \multirow[t]{3}{*}{\entropy} & \name & \q{5.0}{\thousand} & $\float[1]{179.18}_{\pm\float[1]{0.12}}$ & $\float[1]{118.2}_{\pm\float[1]{1.26}}$ & $\float[1]{2.66}_{\pm\float[1]{0.1}}$ & \q{5.0}{\thousand} & $\float[1]{179.18}_{\pm\float[1]{0.12}}$ & $\float[1]{118.2}_{\pm\float[1]{1.26}}$ & $\float[1]{2.66}_{\pm\float[1]{0.1}}$ \\
        &  &  & \randomsubset & \q{5.0}{\thousand} & $\float[1]{179.21}_{\pm\float[1]{0.39}}$ & $\float[1]{117.99}_{\pm\float[1]{4.03}}$ & $\float[1]{21.53}_{\pm\float[1]{0.1}}$ & \q{5.0}{\thousand} & $\float[1]{179.21}_{\pm\float[1]{0.39}}$ & $\float[1]{117.99}_{\pm\float[1]{4.03}}$ & $\float[1]{21.53}_{\pm\float[1]{0.1}}$ \\
        &  &  & \seals & \q{5.0}{\thousand} & $\float[1]{179.55}_{\pm\float[1]{0.06}}$ & $\float[1]{120.8}_{\pm\float[1]{0.79}}$ & $\float[1]{62.16}_{\pm\float[1]{1.25}}$ & \q{5.0}{\thousand} & $\float[1]{179.55}_{\pm\float[1]{0.06}}$ & $\float[1]{120.8}_{\pm\float[1]{0.79}}$ & $\float[1]{62.16}_{\pm\float[1]{1.25}}$ \\
        \cmidrule{3-12}
        &  & \multirow[t]{3}{*}{\ftbertkm} & \name & \q{5.0}{\thousand} & $\float[1]{178.16}_{\pm\float[1]{0.51}}$ & $\float[1]{110.27}_{\pm\float[1]{5.36}}$ & $\float[1]{2.86}_{\pm\float[1]{0.09}}$ & \q{5.0}{\thousand} & $\float[1]{178.16}_{\pm\float[1]{0.51}}$ & $\float[1]{110.27}_{\pm\float[1]{5.36}}$ & $\float[1]{2.86}_{\pm\float[1]{0.09}}$ \\
        &  &  & \randomsubset & \q{5.0}{\thousand} & $\float[1]{176.59}_{\pm\float[1]{0.29}}$ & $\float[1]{95.87}_{\pm\float[1]{3.38}}$ & $\float[1]{25.06}_{\pm\float[1]{0.23}}$ & \q{5.0}{\thousand} & $\float[1]{176.59}_{\pm\float[1]{0.29}}$ & $\float[1]{95.87}_{\pm\float[1]{3.38}}$ & $\float[1]{25.06}_{\pm\float[1]{0.23}}$ \\
        &  &  & \seals & \q{5.0}{\thousand} & $\float[1]{176.52}_{\pm\float[1]{0.64}}$ & $\float[1]{96.22}_{\pm\float[1]{6.87}}$ & $\float[1]{84.36}_{\pm\float[1]{4.63}}$ & \q{5.0}{\thousand} & $\float[1]{176.52}_{\pm\float[1]{0.64}}$ & $\float[1]{96.22}_{\pm\float[1]{6.87}}$ & $\float[1]{84.36}_{\pm\float[1]{4.63}}$ \\
        \cmidrule{3-12}
        &  & \random & \noop & \q{5.0}{\thousand} & $\float[1]{172.44}_{\pm\float[1]{1.16}}$ & $\float[1]{59.35}_{\pm\float[1]{12.86}}$ & $\float[1]{0.03}_{\pm\float[1]{0.0}}$ & \q{5.0}{\thousand} & $\float[1]{172.44}_{\pm\float[1]{1.16}}$ & $\float[1]{59.35}_{\pm\float[1]{12.86}}$ & $\float[1]{0.03}_{\pm\float[1]{0.0}}$ \\
        \cmidrule{2-12}
        & \multirow[t]{3}{*}{\berttiny} & \multirow[t]{3}{*}{\entropy} & \name & \q{5.0}{\thousand} & $\float[1]{172.55}_{\pm\float[1]{1.2}}$ & $\float[1]{54.28}_{\pm\float[1]{14.44}}$ & $\float[1]{1.38}_{\pm\float[1]{0.03}}$ & \q{5.0}{\thousand} & $\float[1]{172.55}_{\pm\float[1]{1.2}}$ & $\float[1]{54.28}_{\pm\float[1]{14.44}}$ & $\float[1]{1.38}_{\pm\float[1]{0.03}}$ \\
        &  &  & \randomsubset & \q{5.0}{\thousand} & $\float[1]{168.31}_{\pm\float[1]{0.73}}$ & $\float[1]{4.02}_{\pm\float[1]{9.61}}$ & $\float[1]{7.53}_{\pm\float[1]{0.12}}$ & \q{5.0}{\thousand} & $\float[1]{168.31}_{\pm\float[1]{0.73}}$ & $\float[1]{4.02}_{\pm\float[1]{9.61}}$ & $\float[1]{7.53}_{\pm\float[1]{0.12}}$ \\
        &  &  & \seals & \q{5.0}{\thousand} & $\float[1]{168.11}_{\pm\float[1]{0.27}}$ & $\float[1]{1.56}_{\pm\float[1]{3.79}}$ & $\float[1]{26.32}_{\pm\float[1]{0.61}}$ & \q{5.0}{\thousand} & $\float[1]{168.11}_{\pm\float[1]{0.27}}$ & $\float[1]{1.56}_{\pm\float[1]{3.79}}$ & $\float[1]{26.32}_{\pm\float[1]{0.61}}$ \\
        \cmidrule{2-12}
        & \multirow[t]{3}{*}{\debertabase} & \multirow[t]{3}{*}{\entropy} & \name & \q{3.5}{\thousand} & $\float[1]{125.32}_{\pm\float[1]{0.3}}$ & $\float[1]{75.79}_{\pm\float[1]{2.36}}$ & $\float[1]{2.31}_{\pm\float[1]{0.03}}$ & \q{2.9}{\thousand} \cellcolor{gray!30} & $\float[1]{102.42}_{\pm\float[1]{0.53}}$ & $\float[1]{60.75}_{\pm\float[1]{3.58}}$ & $\float[1]{1.9}_{\pm\float[1]{0.1}}$ \\
        &  &  & \randomsubset & \q{3.1}{\thousand} & $\float[1]{110.79}_{\pm\float[1]{0.07}}$ & $\float[1]{65.97}_{\pm\float[1]{1.05}}$ & $\float[1]{21.03}_{\pm\float[1]{0.28}}$ & \q{2.9}{\thousand} \cellcolor{gray!30} & $\float[1]{102.65}_{\pm\float[1]{0.26}}$ & $\float[1]{61.52}_{\pm\float[1]{2.72}}$ & $\float[1]{19.42}_{\pm\float[1]{0.18}}$ \\
        &  &  & \seals & \q{2.9}{\thousand} & $\float[1]{102.6}_{\pm\float[1]{0.19}}$ & $\float[1]{60.52}_{\pm\float[1]{1.97}}$ & $\float[1]{40.14}_{\pm\float[1]{0.71}}$ & \q{2.9}{\thousand} & $\float[1]{102.6}_{\pm\float[1]{0.19}}$ & $\float[1]{60.52}_{\pm\float[1]{1.97}}$ & $\float[1]{40.14}_{\pm\float[1]{0.71}}$ \\
        \cmidrule{2-12}
        & \multirow[t]{3}{*}{\gpt} & \multirow[t]{3}{*}{\entropy} & \name & \q{5.0}{\thousand} & $\float[1]{168.0}_{\pm\float[1]{0.0}}$ & $\float[1]{0.0}_{\pm\float[1]{0.0}}$ & $\float[1]{2.62}_{\pm\float[1]{0.02}}$ & \q{4.7}{\thousand} \cellcolor{gray!30} & $\float[1]{158.57}_{\pm\float[1]{0.0}}$ & $\float[1]{0.0}_{\pm\float[1]{0.0}}$ & $\float[1]{2.46}_{\pm\float[1]{0.02}}$ \\
        &  &  & \randomsubset & \q{5.0}{\thousand} & $\float[1]{168.0}_{\pm\float[1]{0.0}}$ & $\float[1]{0.0}_{\pm\float[1]{0.0}}$ & $\float[1]{20.04}_{\pm\float[1]{0.1}}$ & \q{4.7}{\thousand} \cellcolor{gray!30} & $\float[1]{158.57}_{\pm\float[1]{0.0}}$ & $\float[1]{0.0}_{\pm\float[1]{0.0}}$ & $\float[1]{18.99}_{\pm\float[1]{0.11}}$ \\
        &  &  & \seals & \q{4.7}{\thousand} & $\float[1]{158.57}_{\pm\float[1]{0.0}}$ & $\float[1]{0.0}_{\pm\float[1]{0.0}}$ & $\float[1]{73.41}_{\pm\float[1]{0.65}}$ & \q{4.7}{\thousand} & $\float[1]{158.57}_{\pm\float[1]{0.0}}$ & $\float[1]{0.0}_{\pm\float[1]{0.0}}$ & $\float[1]{73.41}_{\pm\float[1]{0.65}}$ \\
        \cmidrule{2-12}
        & \multirow[t]{3}{*}{\tf} & \multirow[t]{3}{*}{\entropy} & \name & \q{3.7}{\thousand} & $\float[1]{131.58}_{\pm\float[1]{0.18}}$ & $\float[1]{77.94}_{\pm\float[1]{2.04}}$ & $\float[1]{4.04}_{\pm\float[1]{0.03}}$ & \q{3.3}{\thousand} \cellcolor{gray!30} & $\float[1]{116.21}_{\pm\float[1]{0.27}}$ & $\float[1]{69.09}_{\pm\float[1]{2.38}}$ & $\float[1]{3.54}_{\pm\float[1]{0.05}}$ \\
        &  &  & \randomsubset & \q{3.6}{\thousand} & $\float[1]{128.25}_{\pm\float[1]{0.13}}$ & $\float[1]{69.47}_{\pm\float[1]{1.27}}$ & $\float[1]{37.54}_{\pm\float[1]{1.55}}$ & \q{3.3}{\thousand} \cellcolor{gray!30} & $\float[1]{115.42}_{\pm\float[1]{0.26}}$ & $\float[1]{61.17}_{\pm\float[1]{2.34}}$ & $\float[1]{35.18}_{\pm\float[1]{0.19}}$ \\
        &  &  & \seals & \q{3.3}{\thousand} & $\float[1]{115.2}_{\pm\float[1]{0.25}}$ & $\float[1]{58.88}_{\pm\float[1]{2.98}}$ & $\float[1]{96.32}_{\pm\float[1]{6.51}}$ & \q{3.3}{\thousand} & $\float[1]{115.2}_{\pm\float[1]{0.25}}$ & $\float[1]{58.88}_{\pm\float[1]{2.98}}$ & $\float[1]{96.32}_{\pm\float[1]{6.51}}$ \\
        \midrule\midrule
        \multirow[t]{29}{*}{\amazonagri} & \multirow[t]{3}{*}{\albertbase} & \multirow[t]{3}{*}{\entropy} & \name & \q{3.6}{\thousand} & $\float[1]{134.18}_{\pm\float[1]{0.72}}$ & $\float[1]{46.65}_{\pm\float[1]{2.93}}$ & $\float[1]{4.87}_{\pm\float[1]{0.09}}$ & \q{3.1}{\thousand} \cellcolor{gray!30} & $\float[1]{115.88}_{\pm\float[1]{0.86}}$ & $\float[1]{43.46}_{\pm\float[1]{1.41}}$ & $\float[1]{4.13}_{\pm\float[1]{0.09}}$ \\
        &  &  & \randomsubset & \q{3.4}{\thousand} & $\float[1]{129.56}_{\pm\float[1]{0.5}}$ & $\float[1]{27.51}_{\pm\float[1]{16.68}}$ & $\float[1]{31.18}_{\pm\float[1]{0.02}}$ & \q{3.1}{\thousand} \cellcolor{gray!30} & $\float[1]{116.02}_{\pm\float[1]{0.44}}$ & $\float[1]{25.52}_{\pm\float[1]{13.42}}$ & $\float[1]{27.94}_{\pm\float[1]{0.04}}$ \\
        &  &  & \seals & \q{3.1}{\thousand} & $\float[1]{116.4}_{\pm\float[1]{0.28}}$ & $\float[1]{45.44}_{\pm\float[1]{2.15}}$ & $\float[1]{93.09}_{\pm\float[1]{1.15}}$ & \q{3.1}{\thousand} & $\float[1]{116.4}_{\pm\float[1]{0.28}}$ & $\float[1]{45.44}_{\pm\float[1]{2.15}}$ & $\float[1]{93.09}_{\pm\float[1]{1.15}}$ \\
        \cmidrule{2-12}
        & \multirow[t]{13}{*}{\bertbase} & \multirow[t]{4}{*}{\badge} & \name & \q{3.7}{\thousand} & $\float[1]{142.38}_{\pm\float[1]{0.06}}$ & $\float[1]{76.33}_{\pm\float[1]{3.25}}$ & $\float[1]{4.52}_{\pm\float[1]{0.09}}$ & \q{3.1}{\thousand} \cellcolor{gray!30} & $\float[1]{119.77}_{\pm\float[1]{0.06}}$ & $\float[1]{63.54}_{\pm\float[1]{2.79}}$ & $\float[1]{3.76}_{\pm\float[1]{0.09}}$ \\
        &  &  & \noop & \integer{425} & $\float[1]{12.67}_{\pm\float[1]{0.03}}$ & $\float[1]{2.59}_{\pm\float[1]{1.03}}$ &\q{6}{\hour}& \integer{425} & $\float[1]{12.67}_{\pm\float[1]{0.03}}$ & $\float[1]{2.59}_{\pm\float[1]{1.03}}$ &\q{6}{\hour}\\
        &  &  & \randomsubset & \q{3.6}{\thousand} & $\float[1]{138.17}_{\pm\float[1]{0.05}}$ & $\float[1]{60.83}_{\pm\float[1]{1.97}}$ & $\float[1]{30.53}_{\pm\float[1]{0.42}}$ & \q{3.1}{\thousand} \cellcolor{gray!30} & $\float[1]{119.52}_{\pm\float[1]{0.08}}$ & $\float[1]{50.61}_{\pm\float[1]{3.25}}$ & $\float[1]{26.27}_{\pm\float[1]{0.42}}$ \\
        &  &  & \seals & \q{3.2}{\thousand} & $\float[1]{121.75}_{\pm\float[1]{0.01}}$ & $\float[1]{64.42}_{\pm\float[1]{0.47}}$ & $\float[1]{91.52}_{\pm\float[1]{3.42}}$ & \q{3.1}{\thousand} \cellcolor{gray!30} & $\float[1]{119.78}_{\pm\float[1]{0.02}}$ & $\float[1]{63.08}_{\pm\float[1]{0.53}}$ & $\float[1]{89.16}_{\pm\float[1]{3.8}}$ \\
        \cmidrule{3-12}
        &  & \multirow[t]{4}{*}{\entropy} & \name & \q{3.7}{\thousand} & $\float[1]{142.42}_{\pm\float[1]{0.04}}$ & $\float[1]{78.05}_{\pm\float[1]{1.0}}$ & $\float[1]{4.4}_{\pm\float[1]{0.05}}$ & \q{3.1}{\thousand} \cellcolor{gray!30} & $\float[1]{119.79}_{\pm\float[1]{0.05}}$ & $\float[1]{64.87}_{\pm\float[1]{0.89}}$ & $\float[1]{3.64}_{\pm\float[1]{0.06}}$ \\
        &  &  & \noop & \integer{475} & $\float[1]{14.62}_{\pm\float[1]{0.0}}$ & $\float[1]{2.9}_{\pm\float[1]{0.0}}$ &\q{6}{\hour}& \integer{475} & $\float[1]{14.62}_{\pm\float[1]{0.0}}$ & $\float[1]{2.9}_{\pm\float[1]{0.0}}$ &\q{6}{\hour}\\
        &  &  & \randomsubset & \q{3.6}{\thousand} & $\float[1]{138.26}_{\pm\float[1]{0.06}}$ & $\float[1]{64.96}_{\pm\float[1]{3.18}}$ & $\float[1]{28.96}_{\pm\float[1]{0.19}}$ & \q{3.1}{\thousand} \cellcolor{gray!30} & $\float[1]{119.59}_{\pm\float[1]{0.05}}$ & $\float[1]{54.26}_{\pm\float[1]{1.95}}$ & $\float[1]{25.1}_{\pm\float[1]{0.13}}$ \\
        &  &  & \seals & \q{3.2}{\thousand} & $\float[1]{123.72}_{\pm\float[1]{0.02}}$ & $\float[1]{65.59}_{\pm\float[1]{0.78}}$ & $\float[1]{85.37}_{\pm\float[1]{1.28}}$ & \q{3.1}{\thousand} \cellcolor{gray!30} & $\float[1]{119.79}_{\pm\float[1]{0.03}}$ & $\float[1]{63.67}_{\pm\float[1]{1.44}}$ & $\float[1]{81.32}_{\pm\float[1]{2.36}}$ \\
        \cmidrule{3-12}
        &  & \multirow[t]{4}{*}{\ftbertkm} & \name & \q{3.8}{\thousand} & $\float[1]{143.28}_{\pm\float[1]{0.02}}$ & $\float[1]{73.75}_{\pm\float[1]{1.52}}$ & $\float[1]{4.48}_{\pm\float[1]{0.06}}$ & \q{3.1}{\thousand} \cellcolor{gray!30} & $\float[1]{119.69}_{\pm\float[1]{0.03}}$ & $\float[1]{59.82}_{\pm\float[1]{1.7}}$ & $\float[1]{3.69}_{\pm\float[1]{0.09}}$ \\
        &  &  & \noop & \integer{425} & $\float[1]{12.66}_{\pm\float[1]{0.0}}$ & $\float[1]{1.98}_{\pm\float[1]{0.0}}$ &\q{6}{\hour}& \integer{425} & $\float[1]{12.66}_{\pm\float[1]{0.0}}$ & $\float[1]{1.98}_{\pm\float[1]{0.0}}$ &\q{6}{\hour}\\
        &  &  & \randomsubset & \q{3.6}{\thousand} & $\float[1]{137.96}_{\pm\float[1]{0.14}}$ & $\float[1]{52.02}_{\pm\float[1]{7.97}}$ & $\float[1]{31.6}_{\pm\float[1]{0.27}}$ & \q{3.1}{\thousand} \cellcolor{gray!30} & $\float[1]{119.39}_{\pm\float[1]{0.14}}$ & $\float[1]{45.82}_{\pm\float[1]{6.16}}$ & $\float[1]{27.5}_{\pm\float[1]{0.37}}$ \\
        &  &  & \seals & \q{3.1}{\thousand} & $\float[1]{119.62}_{\pm\float[1]{0.1}}$ & $\float[1]{56.49}_{\pm\float[1]{5.56}}$ & $\float[1]{101.61}_{\pm\float[1]{0.89}}$ & \q{3.1}{\thousand} & $\float[1]{119.62}_{\pm\float[1]{0.1}}$ & $\float[1]{56.49}_{\pm\float[1]{5.56}}$ & $\float[1]{101.61}_{\pm\float[1]{0.89}}$ \\
        \cmidrule{3-12}
        &  & \random & \noop & \q{3.7}{\thousand} & $\float[1]{140.23}_{\pm\float[1]{0.06}}$ & $\float[1]{20.78}_{\pm\float[1]{2.63}}$ & $\float[1]{0.26}_{\pm\float[1]{0.01}}$ & \q{3.1}{\thousand} \cellcolor{gray!30} & $\float[1]{118.8}_{\pm\float[1]{0.05}}$ & $\float[1]{16.96}_{\pm\float[1]{2.41}}$ & $\float[1]{0.22}_{\pm\float[1]{0.01}}$ \\
        \cmidrule{2-12}
        & \multirow[t]{4}{*}{\berttiny} & \multirow[t]{4}{*}{\entropy} & \name & \q{5.0}{\thousand} & $\float[1]{191.7}_{\pm\float[1]{0.4}}$ & $\float[1]{63.68}_{\pm\float[1]{21.29}}$ & $\float[1]{2.51}_{\pm\float[1]{0.15}}$ & \q{3.6}{\thousand} \cellcolor{gray!30} & $\float[1]{136.68}_{\pm\float[1]{0.36}}$ & $\float[1]{33.0}_{\pm\float[1]{19.86}}$ & $\float[1]{1.58}_{\pm\float[1]{0.13}}$ \\
        &  &  & \noop & \integer{350} & $\float[1]{9.72}_{\pm\float[1]{0.0}}$ & $\float[1]{0.0}_{\pm\float[1]{0.0}}$ &\q{6}{\hour}& \integer{350} & $\float[1]{9.72}_{\pm\float[1]{0.0}}$ & $\float[1]{0.0}_{\pm\float[1]{0.0}}$ &\q{6}{\hour}\\
        &  &  & \randomsubset & \q{4.7}{\thousand} & $\float[1]{177.6}_{\pm\float[1]{0.1}}$ & $\float[1]{40.53}_{\pm\float[1]{5.58}}$ & $\float[1]{14.16}_{\pm\float[1]{0.05}}$ & \q{3.6}{\thousand} \cellcolor{gray!30} & $\float[1]{136.55}_{\pm\float[1]{0.09}}$ & $\float[1]{27.51}_{\pm\float[1]{3.77}}$ & $\float[1]{10.9}_{\pm\float[1]{0.05}}$ \\
        &  &  & \seals & \q{3.6}{\thousand} & $\float[1]{136.91}_{\pm\float[1]{0.14}}$ & $\float[1]{48.11}_{\pm\float[1]{8.91}}$ & $\float[1]{39.63}_{\pm\float[1]{1.42}}$ & \q{3.6}{\thousand} & $\float[1]{136.91}_{\pm\float[1]{0.14}}$ & $\float[1]{48.11}_{\pm\float[1]{8.91}}$ & $\float[1]{39.63}_{\pm\float[1]{1.42}}$ \\
        \cmidrule{2-12}
        & \multirow[t]{3}{*}{\debertabase} & \multirow[t]{3}{*}{\entropy} & \name & \q{1.1}{\thousand} & $\float[1]{41.13}_{\pm\float[1]{0.03}}$ & $\float[1]{18.11}_{\pm\float[1]{1.14}}$ & $\float[1]{1.9}_{\pm\float[1]{0.07}}$ & \q{1.1}{\thousand} & $\float[1]{40.15}_{\pm\float[1]{0.03}}$ & $\float[1]{17.58}_{\pm\float[1]{1.17}}$ & $\float[1]{1.85}_{\pm\float[1]{0.07}}$ \\
        &  &  & \randomsubset & \q{1.1}{\thousand} & $\float[1]{40.92}_{\pm\float[1]{0.17}}$ & $\float[1]{5.65}_{\pm\float[1]{12.02}}$ & $\float[1]{13.99}_{\pm\float[1]{0.05}}$ & \q{1.1}{\thousand} & $\float[1]{39.95}_{\pm\float[1]{0.17}}$ & $\float[1]{5.42}_{\pm\float[1]{11.61}}$ & $\float[1]{13.67}_{\pm\float[1]{0.05}}$ \\
        &  &  & \seals & \q{1.1}{\thousand} & $\float[1]{40.16}_{\pm\float[1]{0.02}}$ & $\float[1]{18.0}_{\pm\float[1]{1.49}}$ & $\float[1]{25.69}_{\pm\float[1]{1.35}}$ & \q{1.1}{\thousand} & $\float[1]{40.16}_{\pm\float[1]{0.02}}$ & $\float[1]{18.0}_{\pm\float[1]{1.49}}$ & $\float[1]{25.69}_{\pm\float[1]{1.35}}$ \\
        \cmidrule{2-12}
        & \multirow[t]{3}{*}{\gpt} & \multirow[t]{3}{*}{\entropy} & \name & \q{3.6}{\thousand} & $\float[1]{136.12}_{\pm\float[1]{0.0}}$ & $\float[1]{0.0}_{\pm\float[1]{0.0}}$ & $\float[1]{3.66}_{\pm\float[1]{0.02}}$ & \q{3.2}{\thousand} \cellcolor{gray!30} & $\float[1]{119.59}_{\pm\float[1]{0.0}}$ & $\float[1]{0.0}_{\pm\float[1]{0.0}}$ & $\float[1]{3.67}_{\pm\float[1]{1.0}}$ \\
        &  &  & \randomsubset & \q{3.5}{\thousand} & $\float[1]{131.26}_{\pm\float[1]{0.0}}$ & $\float[1]{0.0}_{\pm\float[1]{0.0}}$ & $\float[1]{33.24}_{\pm\float[1]{0.05}}$ & \q{3.2}{\thousand} \cellcolor{gray!30} & $\float[1]{119.59}_{\pm\float[1]{0.0}}$ & $\float[1]{0.0}_{\pm\float[1]{0.0}}$ & $\float[1]{30.33}_{\pm\float[1]{0.2}}$ \\
        &  &  & \seals & \q{3.2}{\thousand} & $\float[1]{119.59}_{\pm\float[1]{0.0}}$ & $\float[1]{0.0}_{\pm\float[1]{0.0}}$ & $\float[1]{82.14}_{\pm\float[1]{0.44}}$ & \q{3.2}{\thousand} & $\float[1]{119.59}_{\pm\float[1]{0.0}}$ & $\float[1]{0.0}_{\pm\float[1]{0.0}}$ & $\float[1]{82.14}_{\pm\float[1]{0.44}}$ \\
        \cmidrule{2-12}
        & \multirow[t]{3}{*}{\tf} & \multirow[t]{3}{*}{\entropy} & \name & \q{2.3}{\thousand} & $\float[1]{85.99}_{\pm\float[1]{0.05}}$ & $\float[1]{24.08}_{\pm\float[1]{2.55}}$ & $\float[1]{4.32}_{\pm\float[1]{0.13}}$ & \q{2.2}{\thousand} \cellcolor{gray!30} & $\float[1]{82.08}_{\pm\float[1]{0.1}}$ & $\float[1]{23.2}_{\pm\float[1]{5.86}}$ & $\float[1]{4.18}_{\pm\float[1]{0.19}}$ \\
        &  &  & \randomsubset & \q{2.4}{\thousand} & $\float[1]{89.45}_{\pm\float[1]{0.0}}$ & $\float[1]{0.0}_{\pm\float[1]{0.0}}$ & $\float[1]{43.9}_{\pm\float[1]{0.2}}$ & \q{2.2}{\thousand} \cellcolor{gray!30} & $\float[1]{81.67}_{\pm\float[1]{0.0}}$ & $\float[1]{0.0}_{\pm\float[1]{0.2}}$ & $\float[1]{40.15}_{\pm\float[1]{0.18}}$ \\
        &  &  & \seals & \q{2.2}{\thousand} & $\float[1]{81.67}_{\pm\float[1]{0.0}}$ & $\float[1]{0.0}_{\pm\float[1]{0.0}}$ & $\float[1]{106.48}_{\pm\float[1]{1.9}}$ & \q{2.2}{\thousand} & $\float[1]{81.67}_{\pm\float[1]{0.0}}$ & $\float[1]{0.0}_{\pm\float[1]{0.0}}$ & $\float[1]{106.48}_{\pm\float[1]{1.9}}$ \\
        \midrule\midrule
        \multirow[t]{13}{*}{\amazonmulti} & \multirow[t]{13}{*}{\bertbase} & \multirow[t]{4}{*}{\badge} & \name & \q{3.6}{\thousand} & $\float[1]{110.51}_{\pm\float[1]{1.08}}$ & $\float[1]{82.25}_{\pm\float[1]{14.23}}$ & $\float[1]{8.56}_{\pm\float[1]{0.17}}$ & \q{3.1}{\thousand} \cellcolor{gray!30} & $\float[1]{93.76}_{\pm\float[1]{1.87}}$ & $\float[1]{68.11}_{\pm\float[1]{13.47}}$ & $\float[1]{7.19}_{\pm\float[1]{0.23}}$ \\
        &  &  & \noop & \integer{425} & $\float[1]{9.56}_{\pm\float[1]{0.18}}$ & $\float[1]{4.98}_{\pm\float[1]{2.97}}$ &\q{6}{\hour}& \integer{425} & $\float[1]{9.56}_{\pm\float[1]{0.18}}$ & $\float[1]{4.98}_{\pm\float[1]{2.97}}$ &\q{6}{\hour}\\
        &  &  & \randomsubset & \q{3.5}{\thousand} & $\float[1]{105.97}_{\pm\float[1]{1.23}}$ & $\float[1]{76.59}_{\pm\float[1]{14.63}}$ & $\float[1]{30.72}_{\pm\float[1]{0.26}}$ & \q{3.1}{\thousand} \cellcolor{gray!30} & $\float[1]{93.04}_{\pm\float[1]{0.99}}$ & $\float[1]{65.46}_{\pm\float[1]{16.05}}$ & $\float[1]{26.93}_{\pm\float[1]{0.13}}$ \\
        &  &  & \seals & \q{3.1}{\thousand} & $\float[1]{93.22}_{\pm\float[1]{0.76}}$ & $\float[1]{66.58}_{\pm\float[1]{14.26}}$ & $\float[1]{97.39}_{\pm\float[1]{2.28}}$ & \q{3.1}{\thousand} & $\float[1]{93.22}_{\pm\float[1]{0.76}}$ & $\float[1]{66.58}_{\pm\float[1]{14.26}}$ & $\float[1]{97.39}_{\pm\float[1]{2.28}}$ \\
        \cmidrule{3-12}
        &  & \multirow[t]{4}{*}{\entropy} & \name & \q{3.6}{\thousand} & $\float[1]{109.24}_{\pm\float[1]{1.69}}$ & $\float[1]{77.85}_{\pm\float[1]{9.94}}$ & $\float[1]{7.78}_{\pm\float[1]{0.22}}$ & \q{3.1}{\thousand} \cellcolor{gray!30} & $\float[1]{92.66}_{\pm\float[1]{1.72}}$ & $\float[1]{65.27}_{\pm\float[1]{12.36}}$ & $\float[1]{6.67}_{\pm\float[1]{0.4}}$ \\
        &  &  & \noop & \integer{475} & $\float[1]{10.84}_{\pm\float[1]{0.29}}$ & $\float[1]{4.7}_{\pm\float[1]{3.24}}$ &\q{6}{\hour}& \integer{475} & $\float[1]{10.84}_{\pm\float[1]{0.29}}$ & $\float[1]{4.7}_{\pm\float[1]{3.24}}$ &\q{6}{\hour}\\
        &  &  & \randomsubset & \q{3.5}{\thousand} & $\float[1]{106.17}_{\pm\float[1]{0.62}}$ & $\float[1]{74.73}_{\pm\float[1]{14.18}}$ & $\float[1]{26.63}_{\pm\float[1]{0.28}}$ & \q{3.1}{\thousand} \cellcolor{gray!30} & $\float[1]{92.78}_{\pm\float[1]{0.64}}$ & $\float[1]{64.99}_{\pm\float[1]{12.82}}$ & $\float[1]{23.37}_{\pm\float[1]{0.38}}$ \\
        &  &  & \seals & \q{3.2}{\thousand} & $\float[1]{95.33}_{\pm\float[1]{1.58}}$ & $\float[1]{69.2}_{\pm\float[1]{14.71}}$ & $\float[1]{83.82}_{\pm\float[1]{2.67}}$ & \q{3.1}{\thousand} \cellcolor{gray!30} & $\float[1]{93.18}_{\pm\float[1]{1.56}}$ & $\float[1]{67.84}_{\pm\float[1]{12.61}}$ & $\float[1]{81.61}_{\pm\float[1]{3.57}}$ \\
        \cmidrule{3-12}
        &  & \multirow[t]{4}{*}{\ftbertkm} & \name & \q{3.6}{\thousand} & $\float[1]{108.81}_{\pm\float[1]{1.11}}$ & $\float[1]{75.62}_{\pm\float[1]{20.68}}$ & $\float[1]{7.8}_{\pm\float[1]{0.26}}$ & \q{3.1}{\thousand} \cellcolor{gray!30} & $\float[1]{91.21}_{\pm\float[1]{2.78}}$ & $\float[1]{62.71}_{\pm\float[1]{21.25}}$ & $\float[1]{6.44}_{\pm\float[1]{0.32}}$ \\
        &  &  & \noop & \integer{400} & $\float[1]{8.62}_{\pm\float[1]{0.26}}$ & $\float[1]{3.54}_{\pm\float[1]{3.13}}$ &\q{6}{\hour}& \integer{400} & $\float[1]{8.62}_{\pm\float[1]{0.26}}$ & $\float[1]{3.54}_{\pm\float[1]{3.13}}$ &\q{6}{\hour}\\
        &  &  & \randomsubset & \q{3.5}{\thousand} & $\float[1]{105.04}_{\pm\float[1]{1.14}}$ & $\float[1]{72.27}_{\pm\float[1]{13.15}}$ & $\float[1]{29.09}_{\pm\float[1]{0.55}}$ & \q{3.1}{\thousand} \cellcolor{gray!30} & $\float[1]{91.84}_{\pm\float[1]{0.66}}$ & $\float[1]{61.9}_{\pm\float[1]{11.45}}$ & $\float[1]{25.78}_{\pm\float[1]{0.6}}$ \\
        &  &  & \seals & \q{3.1}{\thousand} & $\float[1]{94.17}_{\pm\float[1]{1.45}}$ & $\float[1]{67.42}_{\pm\float[1]{18.28}}$ & $\float[1]{88.2}_{\pm\float[1]{1.11}}$ & \q{3.1}{\thousand} & $\float[1]{92.61}_{\pm\float[1]{1.46}}$ & $\float[1]{66.27}_{\pm\float[1]{18.13}}$ & $\float[1]{85.8}_{\pm\float[1]{1.53}}$ \\
        \cmidrule{3-12}
        &  & \random & \noop & \q{3.6}{\thousand} & $\float[1]{104.01}_{\pm\float[1]{3.68}}$ & $\float[1]{63.92}_{\pm\float[1]{36.99}}$ & $\float[1]{0.26}_{\pm\float[1]{0.0}}$ & \q{3.1}{\thousand} \cellcolor{gray!30} & $\float[1]{89.42}_{\pm\float[1]{3.48}}$ & $\float[1]{54.16}_{\pm\float[1]{32.74}}$ & $\float[1]{0.22}_{\pm\float[1]{0.0}}$ \\
        \midrule\midrule
        \multirow[t]{13}{*}{\wikitoxic} & \multirow[t]{13}{*}{\bertbase} & \multirow[t]{4}{*}{\badge} & \name & \q{4.1}{\thousand} & $\float[1]{126.18}_{\pm\float[1]{0.33}}$ & $\float[1]{118.19}_{\pm\float[1]{0.89}}$ & $\float[1]{2.66}_{\pm\float[1]{0.02}}$ & \q{3.5}{\thousand} \cellcolor{gray!30} & $\float[1]{106.64}_{\pm\float[1]{0.66}}$ & $\float[1]{99.37}_{\pm\float[1]{1.46}}$ & $\float[1]{2.21}_{\pm\float[1]{0.03}}$ \\
        &  &  & \noop & \q{2.9}{\thousand} & $\float[1]{85.75}_{\pm\float[1]{0.0}}$ & $\float[1]{77.04}_{\pm\float[1]{0.0}}$ &\q{6}{\hour}& \q{2.9}{\thousand} & $\float[1]{85.75}_{\pm\float[1]{0.0}}$ & $\float[1]{77.04}_{\pm\float[1]{0.0}}$ &\q{6}{\hour}\\
        &  &  & \randomsubset & \q{3.9}{\thousand} & $\float[1]{119.45}_{\pm\float[1]{0.36}}$ & $\float[1]{109.0}_{\pm\float[1]{0.85}}$ & $\float[1]{21.83}_{\pm\float[1]{0.11}}$ & \q{3.5}{\thousand} \cellcolor{gray!30} & $\float[1]{104.78}_{\pm\float[1]{0.71}}$ & $\float[1]{94.63}_{\pm\float[1]{1.51}}$ & $\float[1]{19.25}_{\pm\float[1]{0.08}}$ \\
        &  &  & \seals & \q{3.6}{\thousand} & $\float[1]{110.0}_{\pm\float[1]{0.63}}$ & $\float[1]{101.68}_{\pm\float[1]{1.42}}$ & $\float[1]{67.12}_{\pm\float[1]{0.48}}$ & \q{3.5}{\thousand} \cellcolor{gray!30} & $\float[1]{105.93}_{\pm\float[1]{0.55}}$ & $\float[1]{97.65}_{\pm\float[1]{1.42}}$ & $\float[1]{63.56}_{\pm\float[1]{0.47}}$ \\
        \cmidrule{3-12}
        &  & \multirow[t]{4}{*}{\entropy} & \name & \q{4.2}{\thousand} & $\float[1]{128.07}_{\pm\float[1]{0.55}}$ & $\float[1]{119.0}_{\pm\float[1]{1.17}}$ & $\float[1]{2.55}_{\pm\float[1]{0.0}}$ & \q{3.5}{\thousand} \cellcolor{gray!30} & $\float[1]{106.62}_{\pm\float[1]{0.45}}$ & $\float[1]{99.03}_{\pm\float[1]{1.09}}$ & $\float[1]{2.07}_{\pm\float[1]{0.01}}$ \\
        &  &  & \noop & \q{2.9}{\thousand} & $\float[1]{88.28}_{\pm\float[1]{0.39}}$ & $\float[1]{80.82}_{\pm\float[1]{0.96}}$ &\q{6}{\hour}& \q{2.9}{\thousand} & $\float[1]{88.28}_{\pm\float[1]{0.39}}$ & $\float[1]{80.82}_{\pm\float[1]{0.96}}$ &\q{6}{\hour}\\
        &  &  & \randomsubset & \q{4.0}{\thousand} & $\float[1]{119.36}_{\pm\float[1]{0.24}}$ & $\float[1]{107.36}_{\pm\float[1]{0.75}}$ & $\float[1]{20.39}_{\pm\float[1]{0.84}}$ & \q{3.5}{\thousand} \cellcolor{gray!30} & $\float[1]{104.67}_{\pm\float[1]{0.39}}$ & $\float[1]{94.56}_{\pm\float[1]{1.38}}$ & $\float[1]{17.81}_{\pm\float[1]{0.12}}$ \\
        &  &  & \seals & \q{3.6}{\thousand} & $\float[1]{109.59}_{\pm\float[1]{0.71}}$ & $\float[1]{100.58}_{\pm\float[1]{1.84}}$ & $\float[1]{60.89}_{\pm\float[1]{1.54}}$ & \q{3.5}{\thousand} \cellcolor{gray!30} & $\float[1]{105.88}_{\pm\float[1]{0.46}}$ & $\float[1]{97.59}_{\pm\float[1]{1.27}}$ & $\float[1]{56.95}_{\pm\float[1]{1.12}}$ \\
        \cmidrule{3-12}
        &  & \multirow[t]{4}{*}{\ftbertkm} & \name & \q{4.0}{\thousand} & $\float[1]{122.28}_{\pm\float[1]{0.53}}$ & $\float[1]{113.86}_{\pm\float[1]{1.28}}$ & $\float[1]{2.57}_{\pm\float[1]{0.03}}$ & \q{3.5}{\thousand} \cellcolor{gray!30} & $\float[1]{106.79}_{\pm\float[1]{1.09}}$ & $\float[1]{99.9}_{\pm\float[1]{2.51}}$ & $\float[1]{2.21}_{\pm\float[1]{0.03}}$ \\
        &  &  & \noop & \q{2.5}{\thousand} & $\float[1]{73.53}_{\pm\float[1]{0.81}}$ & $\float[1]{62.03}_{\pm\float[1]{2.95}}$ & $\float[1]{198.15}_{\pm\float[1]{2.23}}$ & \q{2.5}{\thousand} & $\float[1]{73.53}_{\pm\float[1]{0.81}}$ & $\float[1]{62.03}_{\pm\float[1]{2.95}}$ & $\float[1]{198.15}_{\pm\float[1]{2.23}}$ \\
        &  &  & \randomsubset & \q{3.8}{\thousand} & $\float[1]{113.4}_{\pm\float[1]{0.26}}$ & $\float[1]{100.35}_{\pm\float[1]{0.7}}$ & $\float[1]{22.8}_{\pm\float[1]{0.13}}$ & \q{3.5}{\thousand} \cellcolor{gray!30} & $\float[1]{103.2}_{\pm\float[1]{0.36}}$ & $\float[1]{91.28}_{\pm\float[1]{0.79}}$ & $\float[1]{20.84}_{\pm\float[1]{0.17}}$ \\
        &  &  & \seals & \q{3.5}{\thousand} & $\float[1]{104.11}_{\pm\float[1]{0.58}}$ & $\float[1]{92.89}_{\pm\float[1]{2.14}}$ & $\float[1]{75.65}_{\pm\float[1]{1.14}}$ & \q{3.5}{\thousand} & $\float[1]{104.11}_{\pm\float[1]{0.58}}$ & $\float[1]{92.89}_{\pm\float[1]{2.14}}$ & $\float[1]{75.65}_{\pm\float[1]{1.14}}$ \\
        \cmidrule{3-12}
        &  & \random & \noop & \q{4.0}{\thousand} & $\float[1]{108.6}_{\pm\float[1]{0.61}}$ & $\float[1]{72.37}_{\pm\float[1]{2.08}}$ & $\float[1]{0.03}_{\pm\float[1]{0.0}}$ & \q{3.5}{\thousand} \cellcolor{gray!30} & $\float[1]{93.52}_{\pm\float[1]{0.53}}$ & $\float[1]{60.61}_{\pm\float[1]{2.43}}$ & $\float[1]{0.03}_{\pm\float[1]{0.0}}$ \\
        \bottomrule
    \end{xltabular}
    \endgroup
\end{landscape}